\RequirePackage{fix-cm}

\documentclass[smallextended]{svjour3}       
\smartqed  

\usepackage{graphicx}
\usepackage{amsmath,amssymb}
\usepackage{lscape}
\usepackage{rotating}
\usepackage{multirow}
\usepackage{booktabs}
\usepackage{bm}
\usepackage{url}
\usepackage{natbib}

\usepackage{subcaption}
\usepackage{color}

\def\|{\,|\,}

\def\obj{{\bf x}}
\def\eqref#1{Eq~\ref{#1}}

\def\P{{\rm P}}

\def\obj{{\bf x}}

\def\data{\ifmmode \mathcal D\else$\data$\fi}
\def\labels{\ifmmode \mathcal L\else$\labels$\fi}
\def\card{\ifmmode \mathcal X\else$\card$\fi}

\def\k{\ifmmode \|\!\mathcal{Y}\!\|\else$\k$\fi}

\def\train{\ifmmode \mathcal T\else$\train$\fi}
\def\test{\ifmmode \mathcal U\else$\train$\fi}
\def\model{\ifmmode \mathcal M\else$\model$\fi}

\def\ANDE^#1{\mathop{{\rm A}#1{\rm DE}}}

\def\eqref#1{Eq~\ref{#1}}

\def\LR^#1{\ensuremath{\text{LR}^#1}}

\begin{document}

\title{On the Inter-relationships among Drift rate, Forgetting rate, Bias/variance profile and Error}

\author{Nayyar A. Zaidi \and Geoffrey I.\ Webb \and Francois Petitjean \and Germain Forestier}

\institute{Nayyar A. Zaidi, Geoffrey I.\ Webb, Francois Petitjean, Germain Forestier \at
           Faculty of Information Technology, Monash University, Clayton, VIC 3800, Australia \\
           \email{\{firstname.lastname\}@monash.edu}  
}

\date{Received: date / Accepted: date}

\maketitle

\begin{abstract}

\sloppy We propose two general and falsifiable hypotheses about expectations on generalization error when learning in the context of concept drift. One posits that as drift rate increases, the forgetting rate that minimizes generalization error will also increase and vice versa. The other posits that as a learner's forgetting rate  increases, the bias/variance profile that minimizes generalization error will have lower variance and vice versa. These hypotheses lead to the concept of the \emph{sweet path}, a path through the 3-d space of alternative drift rates, forgetting rates and bias/variance profiles on which generalization error will be minimized, such that slow drift is coupled with low forgetting and low bias, while rapid drift is coupled with fast forgetting and low variance. We present experiments that support the existence of such a sweet path. We also demonstrate that simple learners that select appropriate forgetting rates and bias/variance profiles are highly competitive with the state-of-the-art in incremental learners for concept drift on real-world drift problems.
\end{abstract}
\section{Introduction} \label{sec_intro}

The world is dynamic, in a constant state of flux. Traditional learning systems that learn static models from historical data are unable to adjust to \emph{concept drift} --- changes in the distributions from which data are drawn. A growing body of experimental machine learning research investigates incremental learners that seek to adjust models as appropriate when confronted with concept drift \citep{Gaber05,Gama09,Aggarwal09,vzliobaite2010learning,Bifet2011PAKDD,Nguyen2014,AccuracyUpdatedEnsemble,Krempl2014survey,SurveyConceptDriftAdaptation,ditzler2015learning}. This paper seeks to inform this line of research by identifying relationships between types of concept drift and the properties of the learners that will best handle those forms of drift. Specifically, we propose and investigate two hypotheses ---
\begin{enumerate}
\item\label{hyp:recency} \emph{The drift-rate/forgetting-rate nexus}. As the rate of concept drift increases, model accuracy will in general be maximized by increasing forgetting rates, and conversely, as the drift rate decreases, model accuracy will in general be maximized by decreasing forgetting rates. Here \emph{increasing forgetting rates} means focusing on more recent evidence by reducing window sizes or increasing decay rates.  \emph{Decreasing forgetting rates} means focusing on longer term evidence by increasing window sizes or decreasing decay rates. 
\item\label{hyp:bias/variance} \emph{The forgetting-rate/bias-variance-profile nexus}. As forgetting rates increase, model accuracy will in general be maximized by altering the bias/variance profile of a learner to decrease variance, and conversely, as forgetting rates increase, model accuracy will in general be maximized by decreasing bias.
\end{enumerate}

The first of these hypotheses is intuitive.  The faster the world is changing, the less relevance older information will have. In consequence, more aggressive forgetting mechanisms, specifically smaller windows or higher decay rates, will be required to exclude older examples for which the trade-off between providing additional information that is relevant to the current state-of-the-world and that which is misleading is weighted too heavily to the latter.

The second hypothesis derives from the hypothesis that when learning from smaller quantities of data, lower variance learners will maximize accuracy due to their ability to avoid overfitting, whereas when learning from larger datasets lower bias learners will maximize accuracy, due to their ability to model the details present in large data~\citep{BrainWebb99}. The more past data we forget, the smaller the effective data quantity from which we learn and, therefore, with high-forgetting rate, low-variance models are more desirable and low-bias models with low-forgetting rate.

\begin{figure}
\centering
\includegraphics[height=180pt]{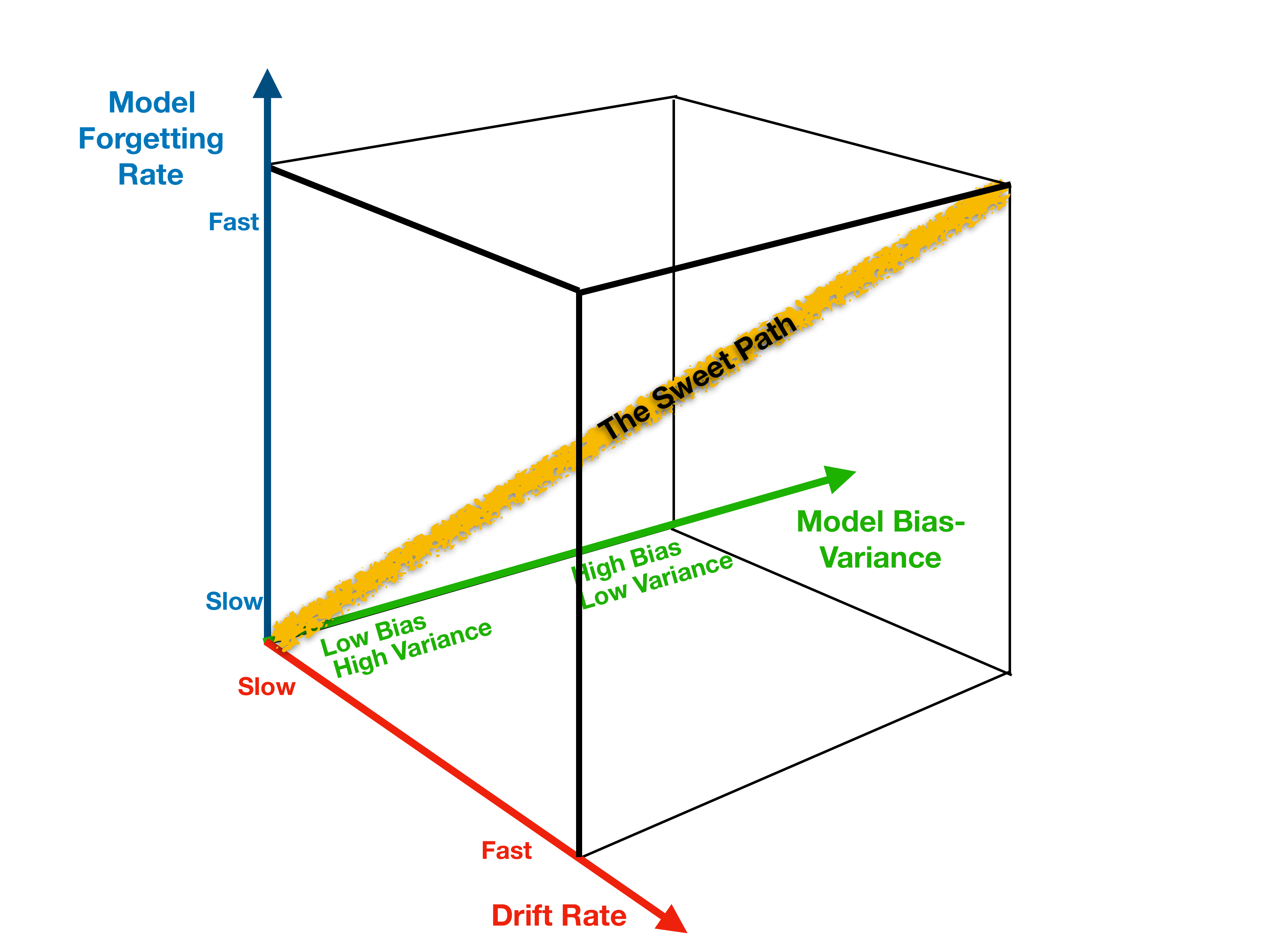}
\caption{\small An illustration of the hypothesized \emph{sweet path}, a four way interaction among drift rate, forgetting rate, learner's bias and variance and expected error.}
\vspace*{-20pt}\label{fig_3wayInteraction3}
\end{figure}
Put together these hypothesized effects imply the \emph{sweet path} for concept drift illustrated in Figure~\ref{fig_3wayInteraction3}, whereby the lowest error for a low drift rate will be achieved by a low bias learner with a low forgetting rate and the lowest error for a high drift rate will be achieved by a low variance learner with a high forgetting rate.

The bulk of this paper (Section~\ref{sec:background} to Section~\ref{sec_experiments}) comprises detailed experiments that investigate the two hypotheses and the hypothesized sweet path. We discuss implications and directions for future research in Section~\ref{sec_conclusions}.

\section{Background}\label{sec:background}

In supervised machine learning we seek to learn a model $\mathcal{M}$ that can predict the value (or probability of each value) $y$ of a target variable $Y$ for an example $x= \langle x_1,\ldots,x_a\rangle$ of an input variable $X=\langle X_1,\ldots,X_a\rangle$.  We learn $\mathcal{M}$ from a training set $T=\{\langle x^1,y^1\rangle, \ldots, \langle x^s,y^s\rangle \}$.  In incremental learning, the training set is presented to the learner as a sequence over a period of time and the learner updates $\mathcal{M}$ in light of each new example or set of examples as it is encountered.

Concept drift occurs when the distribution $P_t(X,Y)$ from which the data are drawn at time $t$ differs from that at subsequent time $u$, $P_u(X,Y)$.\footnote{Note that some papers \citep{SurveyConceptDriftAdaptation} distinguish \emph{real concept drift} in which $P(Y\mid X)$ changes, from virtual concept drift in which $P(X)$ changes. For the purposes of this paper we do not distinguish between these, as the distinction does not appear pertinent.}

We can measure the \emph{magnitude} of drift from time $t$ to $u$, $D(t,u)$, by a measure of distance between the probability distributions $P_t(X,Y)$ and $P_u(X,Y)$ and the \emph{rate} of drift at time $t$ by: 
\begin{equation}
Rate_t=\lim_{n\to\infty}n D\left(t{-}0.5{/}n,t{+}0.5{/}n\right)
\end{equation}
\citep{WebbEtAl16}. The observations in this paper hold for any distance measure that is a metric, such as the Total Variation Distance \citep{levinmarkov}.

Forgetting mechanisms are a standard strategy for dealing with concept drift. The two main forgetting mechanisms are \emph{windowing}, in which a sliding window is maintained containing only the $W$ most recent examples; and \emph{decay} or \emph{weighting}, in which greater weight is placed on more recent examples and lesser weight on older ones~\citep{SurveyConceptDriftAdaptation}.

\section{Experimental setup} \label{sec:setup}
\def\attsubsets{{{\mathcal A}\choose n}}
\def\PANDE^#1{\hat\P_{\textrm{A{#1}DE}}}

\sloppy
{
To explore the {drift-rate/forgetting-rate/bias-variance-profile nexus}, we require an incremental learner that can learn from sliding windows or with decay. Of course, we also require means of varying the learner's bias/variance profile.

For our experiments, we use the semi-naive Bayesian method AnDE \citep{WebbEtAl12}, as it satisfies these requirements.
First, the model has a tuneable parameter $n$ that controls the representation bias and variance. When $n=0$ (in AnDE), one gets a naive Bayes classifier which is highly biased but has low variance.
Higher values of $n$ decrease bias at the cost of an increase in variance and lower values decrease variance at a cost of increased bias. Second, the AnDE model can be represented using counts of observed marginal frequencies, the dimensionality of each of which is controlled by $n$. As described below, these can readily be incrementally updated to reflect a sliding window or incremental decay without need for relearning the entire model.
}

The goal of Bayesian methods is to factorize the joint distribution: $\P(y,\obj)$. The AnDE model factorizes the joint distribution as:
\def\hadjust{\hspace*{-5pt}}
\begin{equation}\label{eq:realande}
\PANDE^n(y, \obj)=\left\{\begin{array}{l@{\quad:\quad}l}
    \displaystyle\sum_{s\in \attsubsets}\delta(x_s)\hat\P(y,x_s)\prod_{i=1}^a\hat\P(x_i\|y,x_s)/\hadjust\sum_{s\in \attsubsets}\delta(x_s)\hadjust&\hadjust\hadjust\displaystyle\sum_{s\in \attsubsets}\delta(x_s)>0\\[20pt]
   \displaystyle \PANDE^{{(n{-}1)}}(y, \obj)&\hadjust\rm otherwise
   \end{array}\right.
\end{equation}
where $\attsubsets$ indicates the set of all size-$n$ subsets of $\{1, \dots a\}$ and $\delta(x_\alpha)$ is a function that is 1 if the training data contains an object with the value $x_\alpha$, otherwise 0.



\subsection{Window-based Adaptation} \label{subsec_wba}
 
A sliding window that supports learning only from the last $W$ data points can be achieved simply with a queue-based data structure.
At each time step $t$, when a new data point $\langle\obj, y\rangle$ arrives:
\begin{itemize}
\item Increment relevant count statistics based on $\langle\obj, y\rangle$
\item Push $\langle\obj, y\rangle$ onto the queue
\item If queue length exceeds $W$
\begin{itemize}
\item ${\langle\tilde\obj, \tilde y\rangle}$ = De-queue. 
\item Decrement relevant count statistics based on ${\langle\tilde\obj, \tilde y\rangle}$
\end{itemize}
\end{itemize}

It can be seen that parameter $W$ controls the forgetting rate. Large $W$ means large windows, hence slow forgetting and small $W$ means small windows, hence fast forgetting.

\subsection{Decay-based Adaptation} \label{subsec_dba}

To support incremental exponential decay, before adding the count statistics of the data point $\obj$ at step $t$, all that is required is that the counts in the count table are decayed. 
For example if $N_{x_i,y}$ denotes the stored count of the number of times attribute $i$ takes value $x_i$ and class attribute takes the value $y$, it is decayed as:
\begin{eqnarray}
N_{x_i,y} = N_{x_i,y} * \exp(-D), \nonumber
\end{eqnarray}
where $D$ is the decay parameter.

Like $W$ in window-based adaptation, it can be seen that parameter $D$ controls the model adaptation rate. 
Large $D$ means large decay, and hence fast model adaptation-rate, and, 
small $D$ means small decay, and hence a slow model adaptation-rate.


\section{Data Generation} \label{sec_datagen}

To test our hypotheses, we require data streams with varying drift rates. 
To this end we create a framework where we can generate synthetic data for which we can systematically manipulate the rate of drift.

\begin{figure}
\centering
\includegraphics[width=80mm,clip,trim={120mm 0 0 0}]{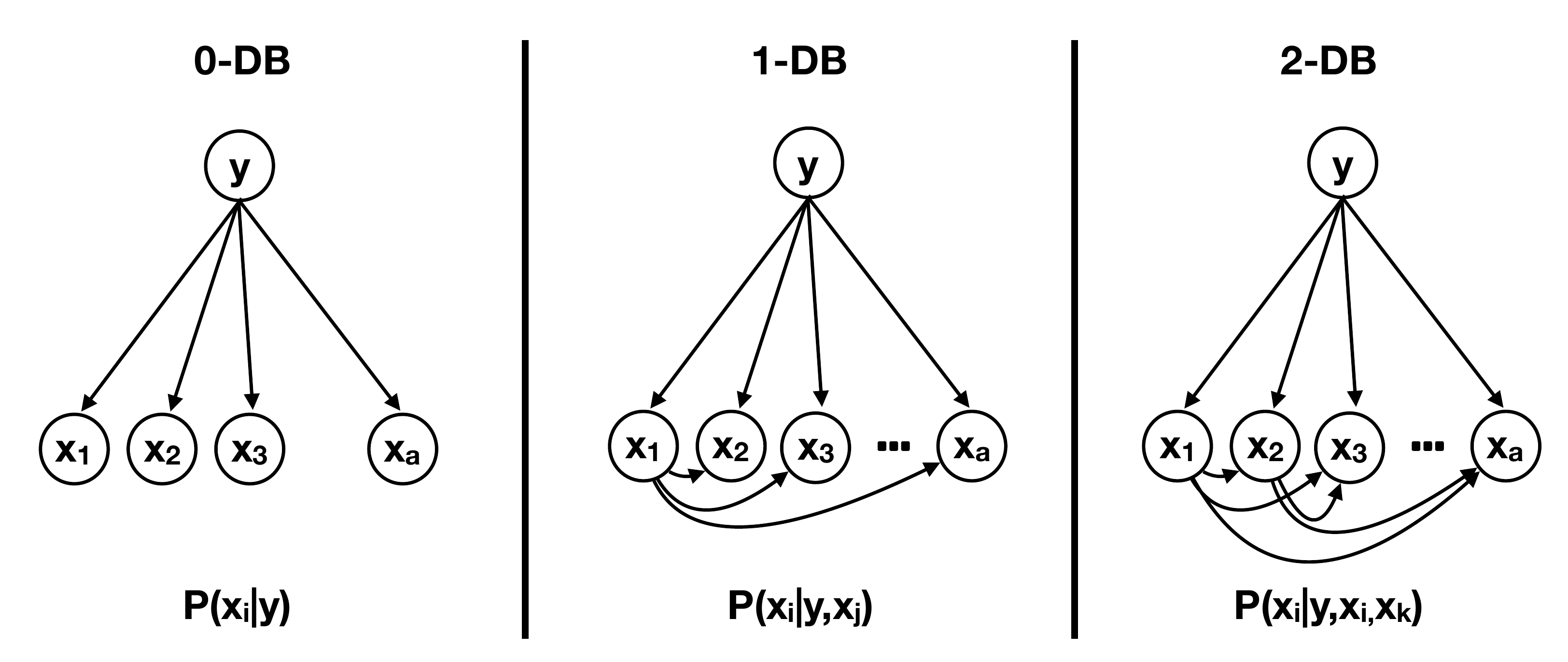}
\caption{\small Simple illustration of the structure  of superparent k-DB classifiers. (Left) 
superparent 1-DB, each variable takes one more parent other than the class, (Right) superparent 2-DB, each variable takes two more parents other than the class.}
\label{fig_kdbDemoFigure}
\end{figure}

We represent the probability distribution using the most common formalism for doing so, a Bayesian network. Because changing the probability distribution at a node may change all of the probability distributions of all its children and their descendant nodes, in order to allow systematic manipulation of the drift rate we minimize the number of parent nodes.  Specifically, we sample from superparent $k$-DB \citep{keogh:labcac} distributions. We show the structure of superparent 
1 and 2-DB in Figure~\ref{fig_kdbDemoFigure}. 
In a superparent $k$-DB model, each attribute $X_i$ other than the first $k$ attributes takes $X_1,\ldots X_k$ and $Y$ as its parents.
In a superparent 1-DB structure every attribute $X_i$ other than $X_1$ takes $Y$ and $X_1$ as its parents and in a superparent 2-DB structure every attribute other than $X_1$ and $X_2$ takes $Y$, $X_1$ and $X_2$ as its parents. These structures are shown in Figure \ref{fig_kdbDemoFigure}.

Specifying such networks is simple -- the standard process is:
\begin{enumerate}
\item Specify the (possibly empty) set $\pi_{X_i}$ of parents for each node ${X_i}$.
\item Fill the Conditional-Probability-Table (CPT) for each node.  This specifies a probability distribution over the values of the attribute for each combination of values of the parents.
\end{enumerate}
Once the network is specified, one can use~\emph{Ancestral Sampling}~\citep{Bishop:2006:PRM:1162264}, to generate data therefrom.

We use the following simple (heuristic) procedure to generate the network:
\begin{itemize}
\item All attributes including the class are binary.
\item Set the initial number of attributes to $100$.
\item We aim for half of these attributes to have three parents, and the remaining half to have two parents in order that the resulting distributions do not too closely fit the biases of a single AnDE learner. Note, however, that as we are creating superparent $k$-DB structures, $X_1$ can only have $Y$ as a parent and that $X_2$ must have $X_1$ and $Y$ as parents. Therefore, in practice, we have $1$ attribute with 1 parent, $49$ attributes with 2 parents and $50$ attributes with 3 parents. 
\item To allow direct control over the rate of drift, we do not allow parent attributes to drift. To maximize diversity, we thus wish to minimize the number of parent attributes, which is why we use superparent $1$-DB and $2$-DB structures instead of more general $k$-DB structures. To this end, for $X_3$ to $X_{50}$, we uniformly at random select either $X_1$ or $X_2$, together with $Y$ as parents. 
For $X_{51}$ to $X_{100}$, all of $Y$, $X_1$ and $X_2$ are assigned as parents. 
\item Once the structure is specified, we randomly initialize the CPTs. Note, we have to fulfill the sum-to-one constraint that $\P(X_i=0\mid \pi_{X_i}) + \P(X_i=1\mid \pi_{X_i}) = 1$. To this end, for each combination of values for the parent attributes, we randomly select a value between 0 and 1 for $\P(X_i=0\mid \pi_{X_i})$, and set $\P(X_i=1\mid \pi_{X_i}) = 1- \P(X_i=0\mid \pi_{X_i})$.
\item Next, we add $100$ more binary attributes which have no parents and hence represent noise. The CPTs for these nodes are also initialized randomly as above. In total, we now have $200$ attributes.
\end{itemize}
To sample from the distribution defined by this network, we:
\begin{itemize}
\item Choose the class $y$ by uniformly at random selecting either 0 or 1.
\item For $i=1$ to $100$ sample $x_i$ from the distribution defined by $P(X_i\mid y, x_1, \ldots, x_{i-1})$.
\end{itemize}

\textbf{Introducing Drift}  
To introduce drift we want to change the CPTs in a controlled manner. We need to strictly control the change  because we need to systematically increase and decrease the rate of drift. To this end we ensure that ---
\begin{itemize}
\item $Y$, $X_1$ and $X_2$ (the three nodes that are parents to the other nodes) do not drift, therefore, their CPTs will not be changed through-out the data generation process. 
\item Drift occurs only after every $T$ steps.
\item Drift only influences $X$\% of the attributes.
\end{itemize}
The first constraint is necessary because changing parent probabilities will indirectly change all the child probabilities in a complex manner that is difficult to manage. The second and third constraints ensure that there is short term directionality in the drift.  Simply randomly drifting every attribute 1/10th of the drift rate every step results in half the steps simply canceling out the previous step for the attribute. However, if only some attributes are drifted at each step and drift occurs only every $T$ steps it ensures that non-trivial drift lasts for a non-trivial period of time. 

Throughout the experiments, we will set $T$ to $10$ and $X$ is set to $50$. That is, half of the attributes are randomly selected after every $10$-th step and are drifted. 

The drifting process is  controlled by a single parameter $\Delta$. The method is designed to ensure that a higher value of $\Delta$ leads to a fast drift, and a smaller value of $\Delta$ will lead to a slow drift. When a node is drifted, $\Delta$ is either added or subtracted from each of its CPT values in a manner to ensure that the values sum to 1.0 and no value exceeds 1.0 or falls below 0.0, hence maintaining a valid probability distribution. 

\section{Experimental Analysis} \label{sec_experiments}

\begin{figure}
\centering

\begin{subfigure}[b]{0.49\textwidth}\includegraphics[width=60mm,height=45mm]{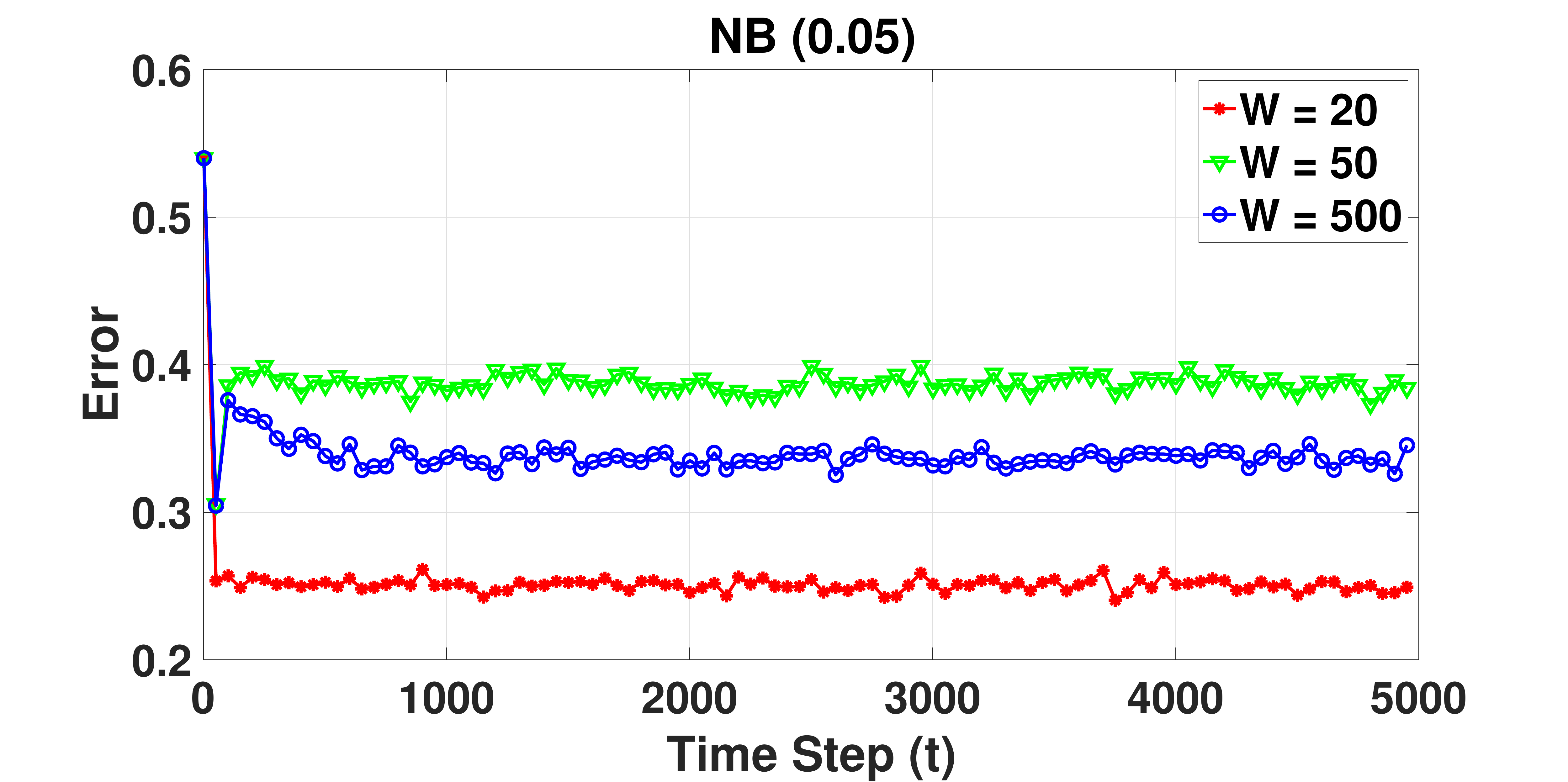} \caption{} \label{fig_wfd_nb} \end{subfigure}
\begin{subfigure}[b]{0.49\textwidth}\includegraphics[width=60mm,height=45mm]{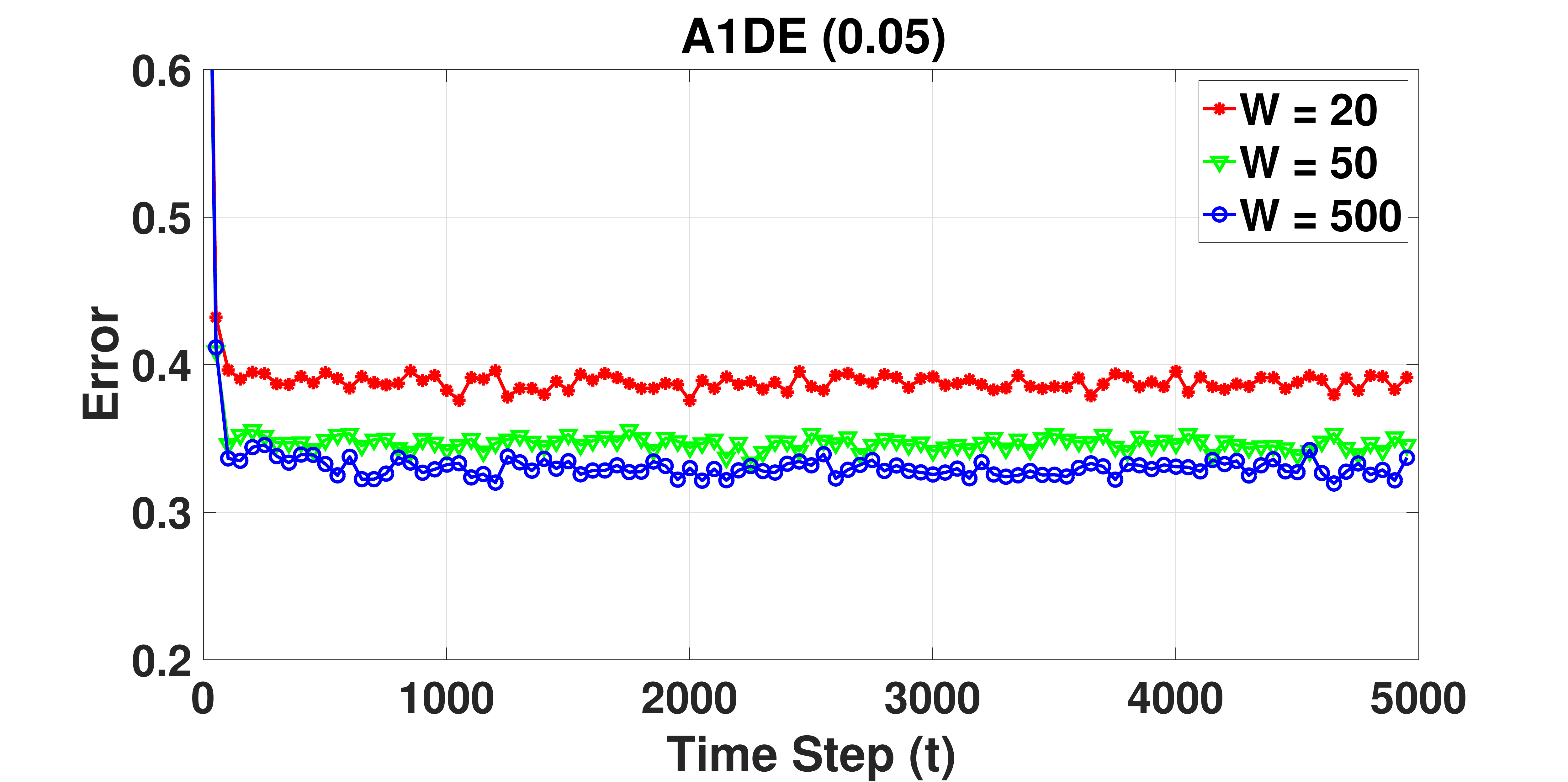} \caption{} \label{fig_wfd_a1de} \end{subfigure}

\begin{subfigure}[b]{0.49\textwidth}\includegraphics[width=60mm,height=45mm]{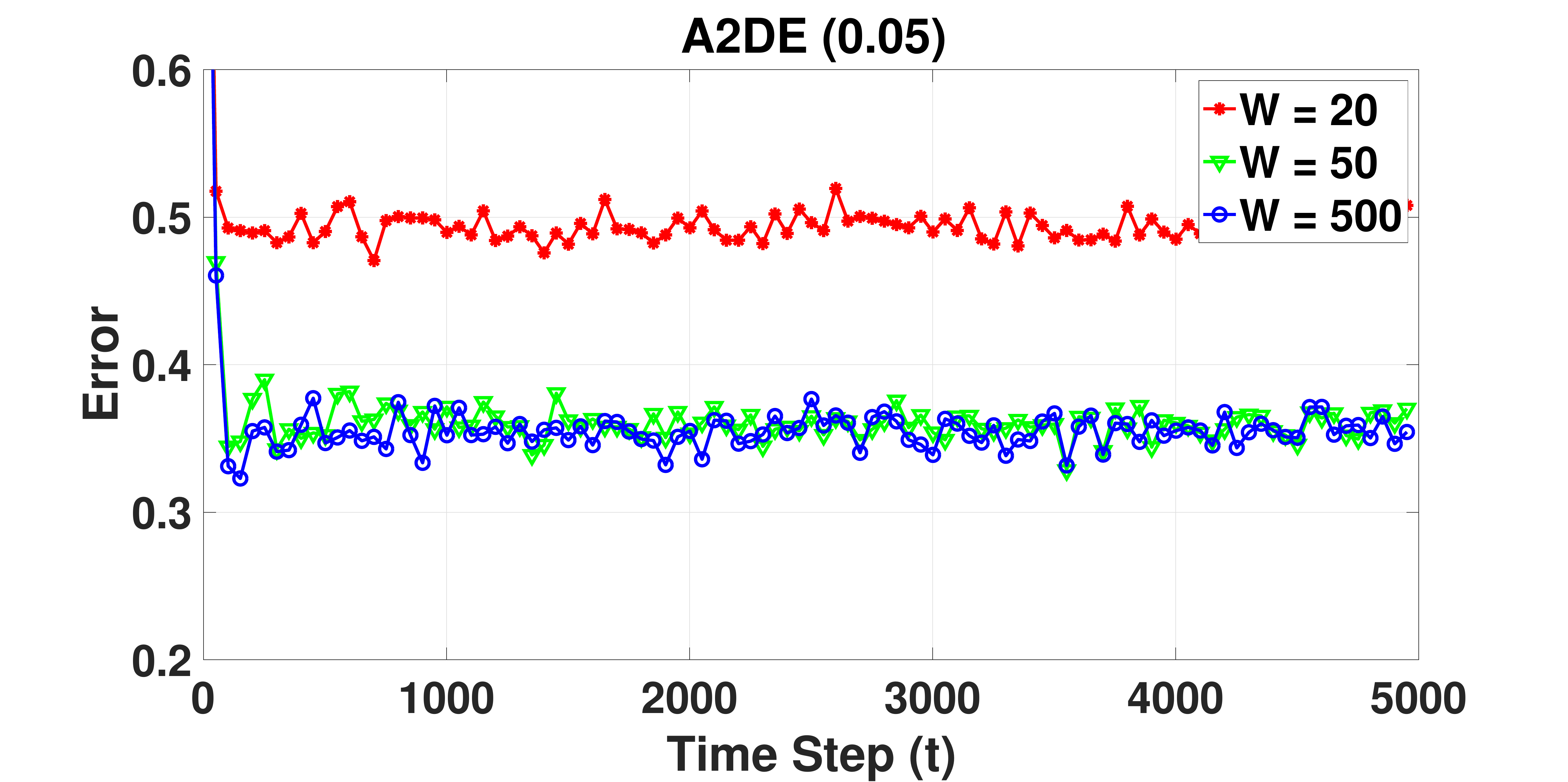} \caption{}  \label{fig_wfd_a2de}\end{subfigure}
\begin{subfigure}[b]{0.49\textwidth}\includegraphics[width=60mm,height=45mm]{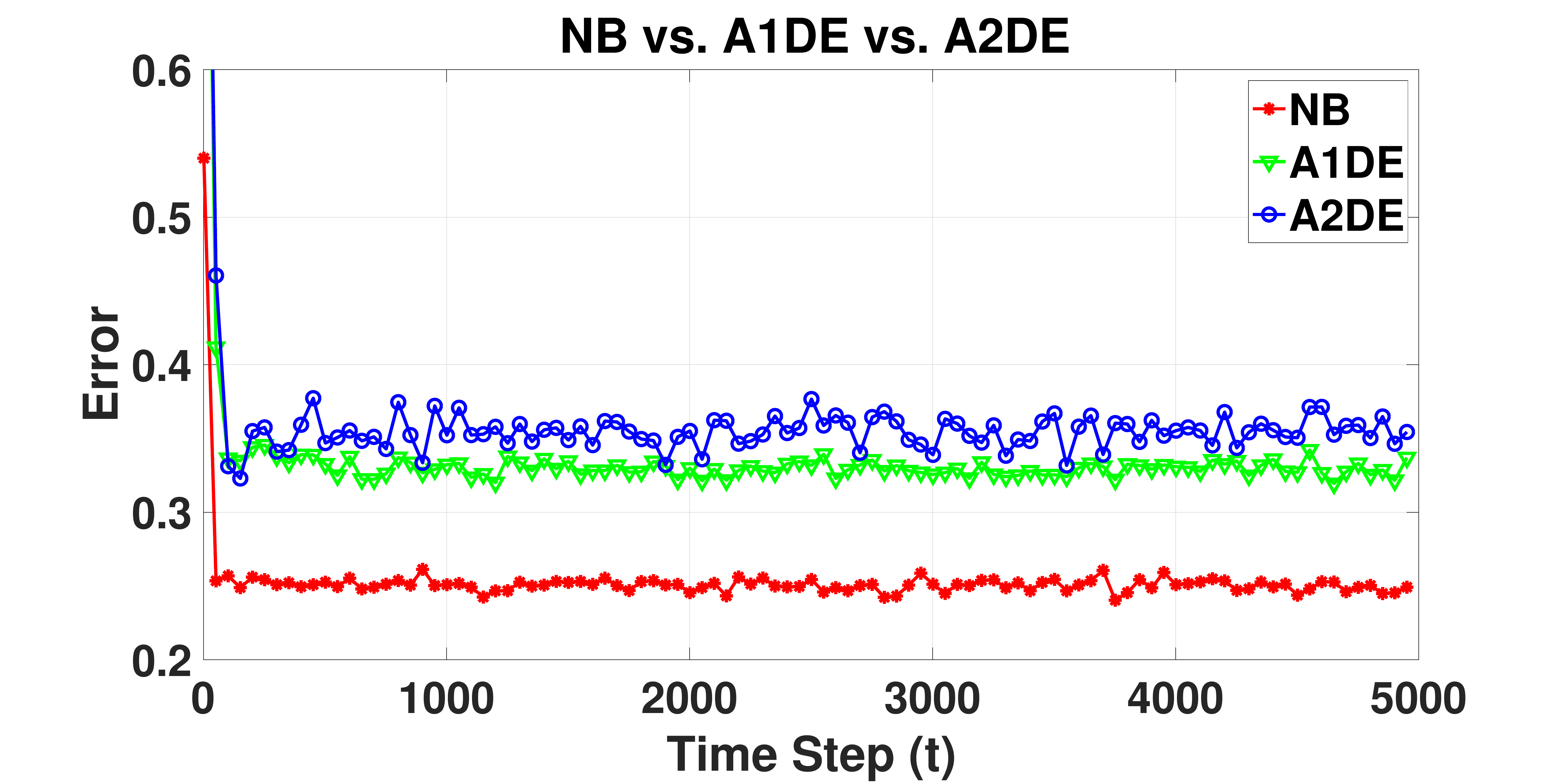} \caption{} \label{fig_wfd_all} \end{subfigure}
\vspace*{-5pt}\caption{\small Windowing with Fast Drift ($\Delta=0.05$) -- Variation in prequential loss of NB (Figure~\ref{fig_wfd_nb}), A1DE (Figure~\ref{fig_wfd_a1de}) and A2DE (Figure~\ref{fig_wfd_a2de}) with window sizes of 20, 50 and 500. Figure~\ref{fig_wfd_all}: Comparison of NB (error = $0.253$), A1DE (error = $0.337$) and A2DE (error = $0.361$) with best window size.}
\vspace*{-10pt}
\label{fig_window_fastDrfit}
\end{figure}
\begin{figure}
\centering

\begin{subfigure}[b]{0.49\textwidth}\includegraphics[width=60mm,height=45mm]{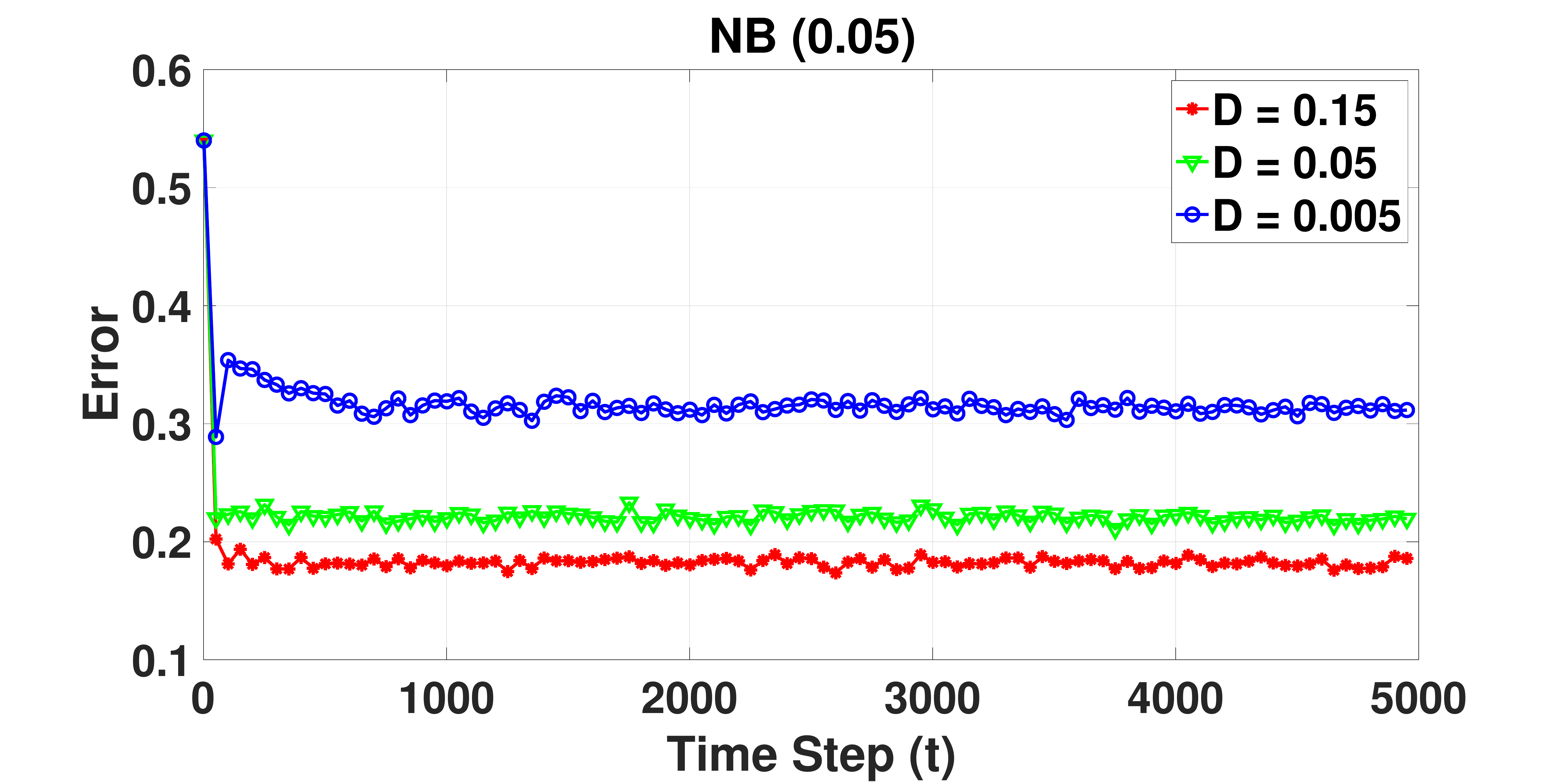} \caption{} \label{fig_dfd_nb} \end{subfigure}
\begin{subfigure}[b]{0.49\textwidth}\includegraphics[width=60mm,height=45mm]{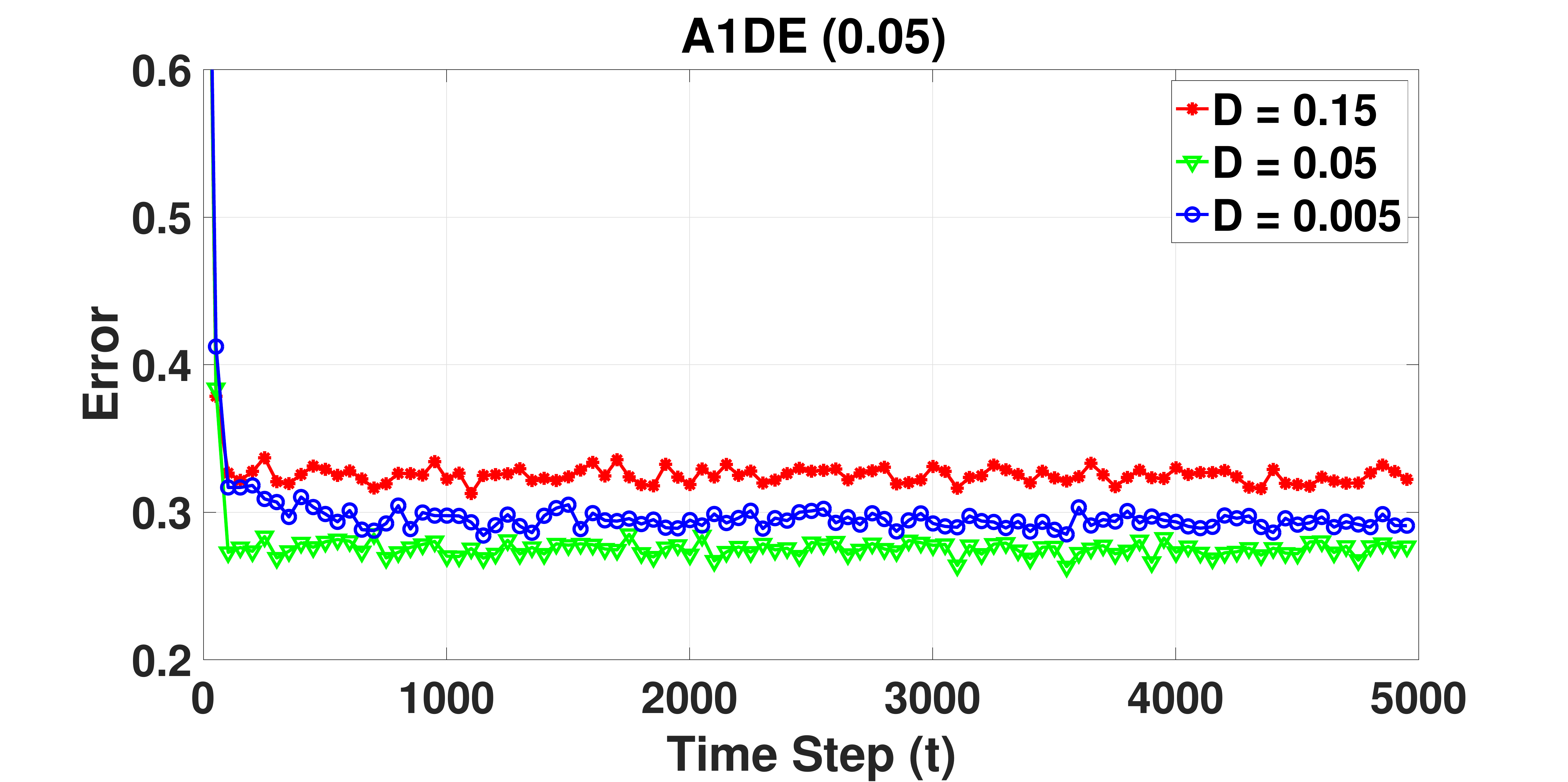} \caption{} \label{fig_dfd_a1de} \end{subfigure}

\begin{subfigure}[b]{0.49\textwidth}\includegraphics[width=60mm,height=45mm]{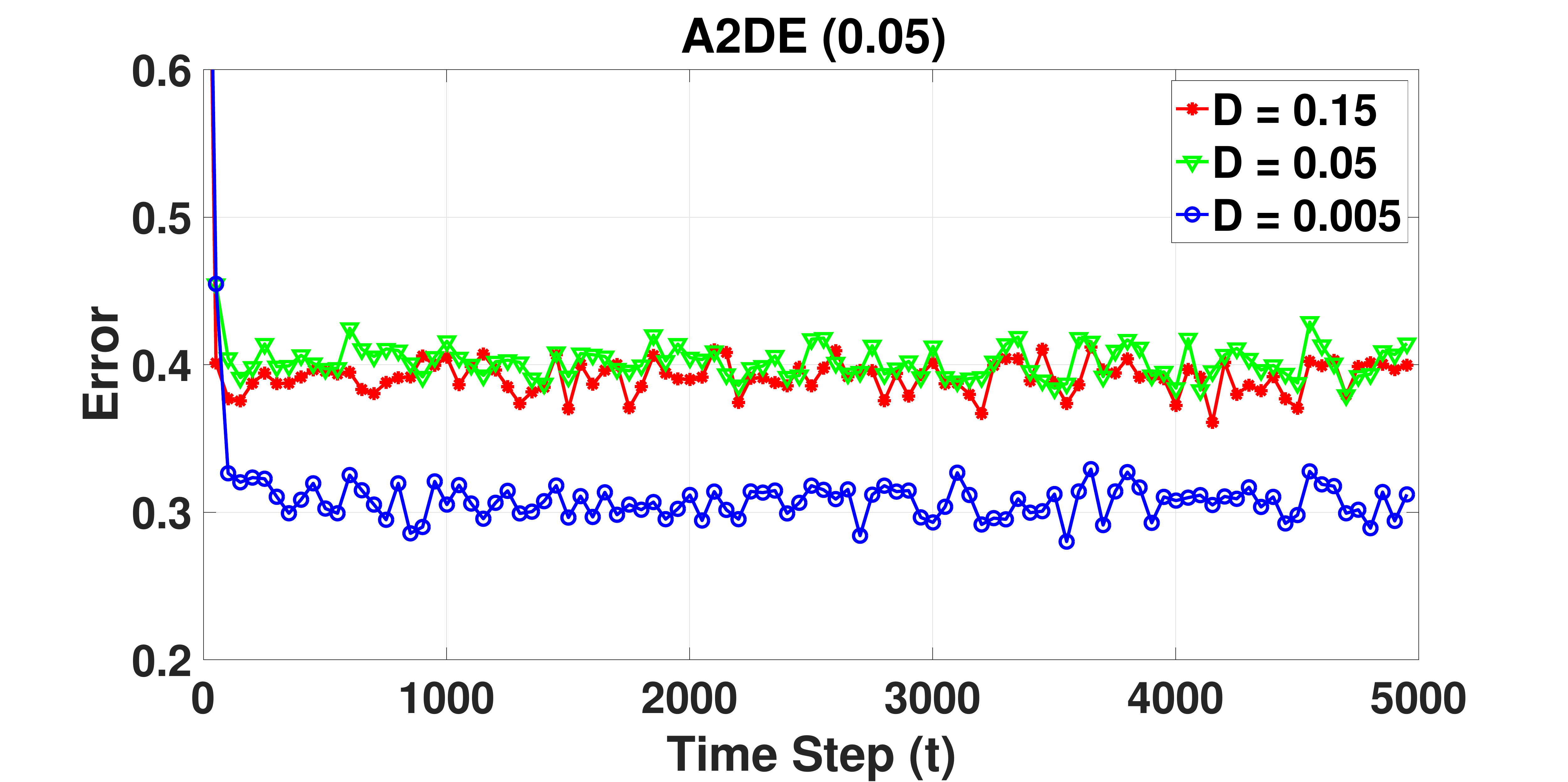} \caption{} \label{fig_dfd_a2de} \end{subfigure}
\begin{subfigure}[b]{0.49\textwidth}\includegraphics[width=60mm,height=45mm]{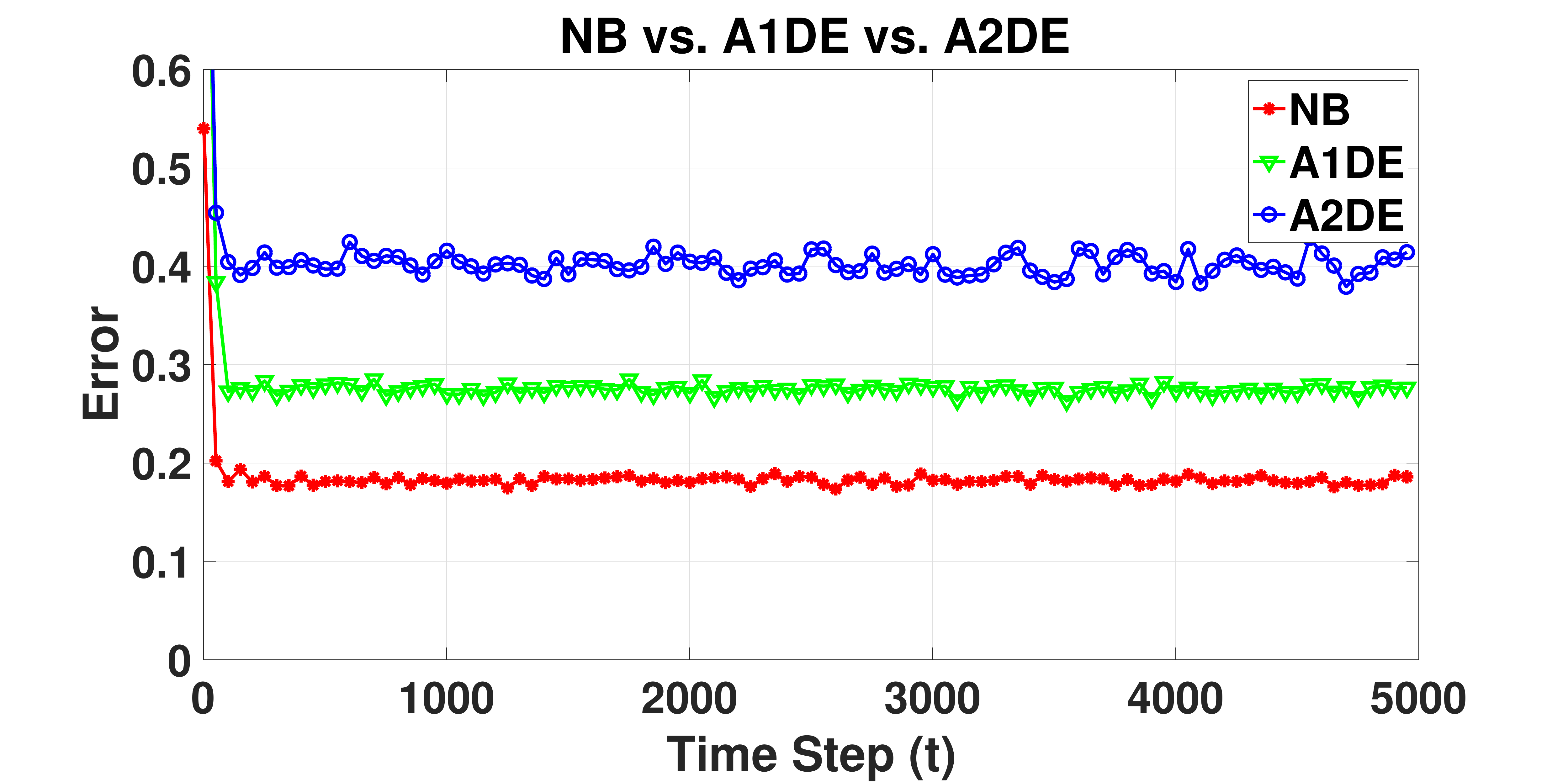} \caption{} \label{fig_dfd_all} \end{subfigure}

\vspace*{-5pt}\caption{\small Decay with Fast Drift ($\Delta=0.05$) -- Variation in prequential loss of NB (Figure~\ref{fig_dfd_nb}), A1DE (Figure~\ref{fig_dfd_a1de}) and A2DE (Figure~\ref{fig_dfd_a2de}) with decay rates of $0.15$, $0.05$ and $0.005$. Figure~\ref{fig_dfd_all}: Comparison of NB (error = $0.186$), A1DE (error = $0.283$) and A2DE (error = $0.408$) with best decay rate.}

\label{fig_decay_fastDrfit}
\end{figure}
We have seen how parameter $n$ controls the bias/variance profile of the model, how parameters $W$ and $D$ control the forgetting rate, and how parameter $\Delta$ controls the rate of the drift. 
In this section, we present experiments that systematically study the interaction among these  factors. 

We use $\Delta=0.05$ to represent fast drift, $\Delta=0.01$ for medium drift and $\Delta=0.0005$ for slow drift.  We select these from a wider range of values explored as exemplars that demonstrate clear differences in outcomes. As an example of the rates not presented, for $\Delta=0.02$ it is less clear which out of our fast and medium forgetting rates provides lower error, as our first hypothesis predicts. 

We generate data streams of 5,000 successive time steps, at each time step drawing one example randomly from the probability distribution for that step and drifting the distribution every 10 steps. 

We use prequential evaluation, whereby at each time step the current model is applied to classify the next example in the data stream and then the example is used to update the model.  We plot the resulting error rates, where each point in the plot is the average error over $50$ successive time steps. We run each experiment 150 times for NB and A1DE and 100 times for A2DE (due to there being insufficient time to complete more runs). We present averages over all runs. 

We first present results for fast drift. Figure~\ref{fig_window_fastDrfit} presents results using windows for forgetting and Figure~\ref{fig_decay_fastDrfit} presents results using decay for forgetting. The average prequential error for the window size or decay rate that achieves the lowest such error is listed for each of the three classifiers (Figures~\ref{fig_wfd_all} and~\ref{fig_dfd_all}).
\clearpage
Here we see that for NB, as predicted, the lowest error is achieved with fast forgetting (window size 20; decay rate $0.15$). 

However, contrary to our expectations, A1DE and A2DE achieve their lowest error with slower forgetting. This is because these models effectively fail in the face of such rapid drift. Recall that A1DE must estimate for every attribute $X_i$ and attribute $X_j$ both $\P(Y, X_i)$ and $\P(X_j\mid Y, X_i)$. A2DE must estimate for every attribute $X_i$, attribute $X_j$ and attribute $X_k$ -- $\P(Y, X_i, X_j)$ and $\P(X_k\mid Y, X_i, X_j)$. The distributions for $Y$, $X_1$ and $X_2$ are not drifting, but all others are drifting at a rapid rate.  Larger window sizes allow these classifiers to produce more accurate estimates of the unvarying probabilities, $\P(Y, X_1)$, $\P(Y, X_2)$, $\P(X_1\mid Y, X_2)$, $\P(X_2\mid Y, X_1)$ and $\P(Y, X_1, X_2)$, whereas no window size provides accurate estimates of the remaining probabilities because either they are too small to provide accurate estimates or the distributions change too much over the duration of the window for the estimate to be accurate.  Thus, the relative error is dominated by the ability to accurately estimate the invariant probabilities and the performance approximates learning from a stationary distribution when the majority of attributes are noise attributes.

\begin{figure}
\centering

\begin{subfigure}[b]{0.49\textwidth}\includegraphics[width=60mm,height=45mm]{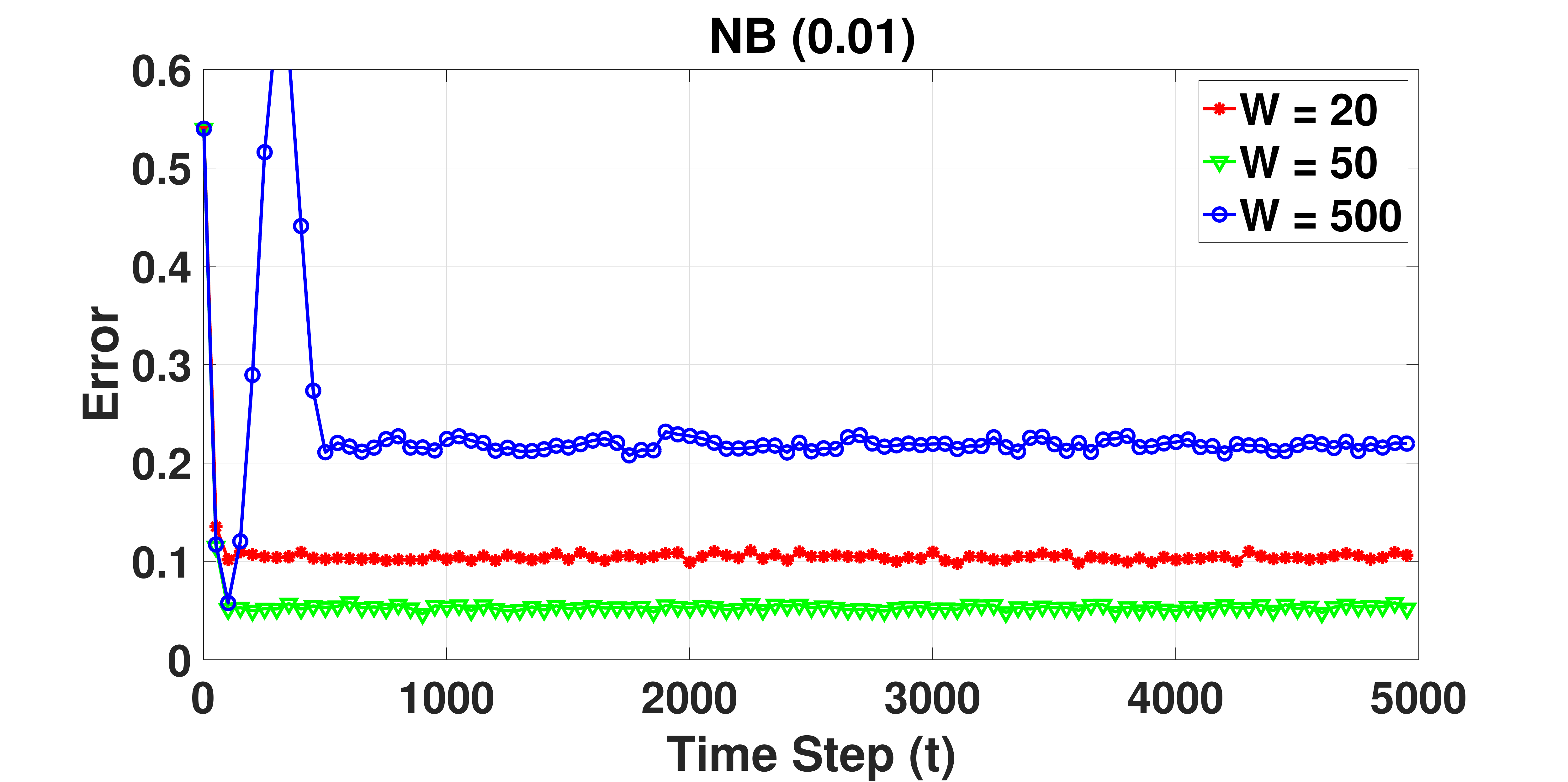}  \caption{} \label{fig_wmd_nb} \end{subfigure}
\begin{subfigure}[b]{0.49\textwidth}\includegraphics[width=60mm,height=45mm]{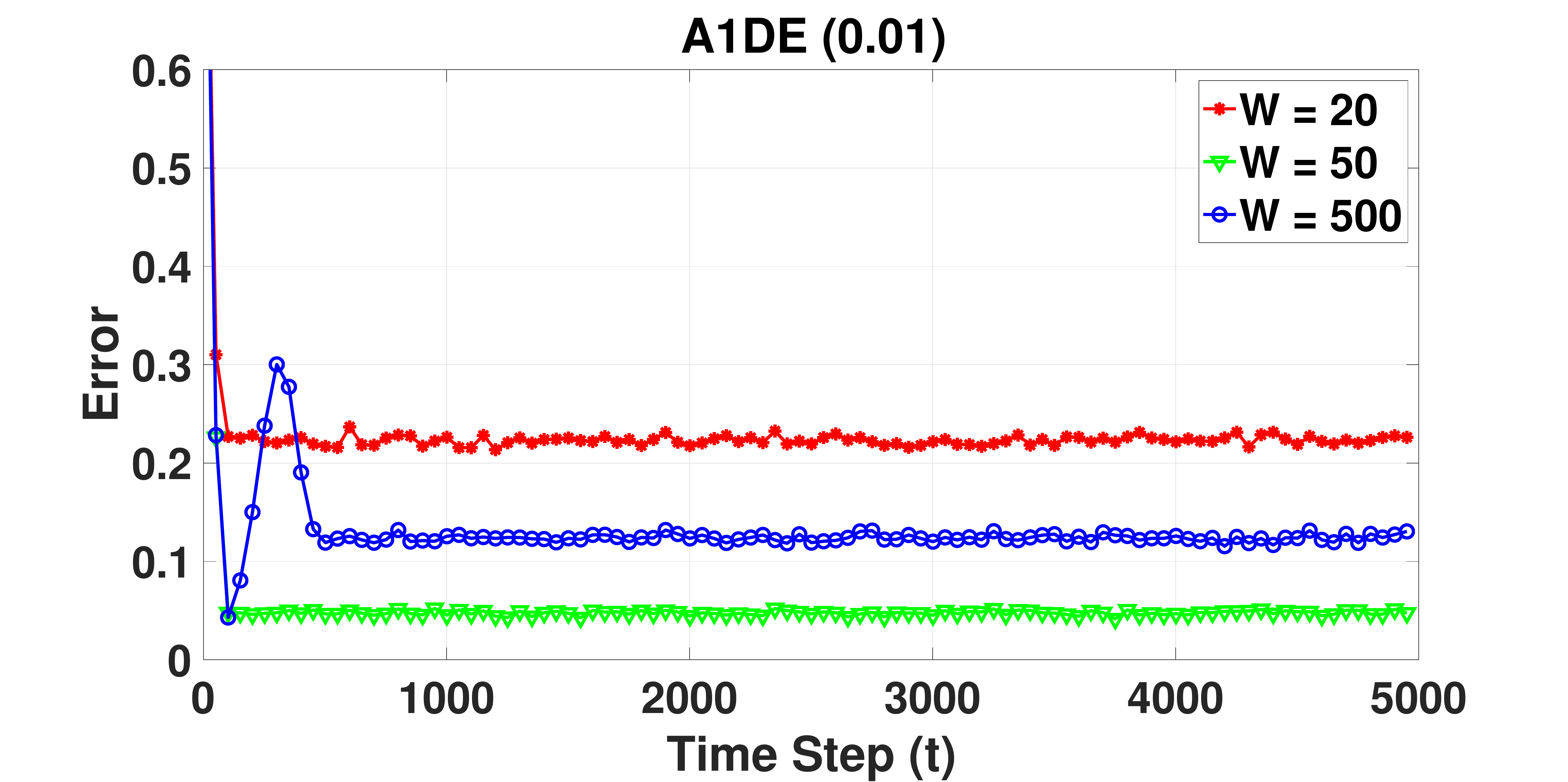} \caption{} \label{fig_wmd_a1de} \end{subfigure}

\begin{subfigure}[b]{0.49\textwidth}\includegraphics[width=60mm,height=45mm]{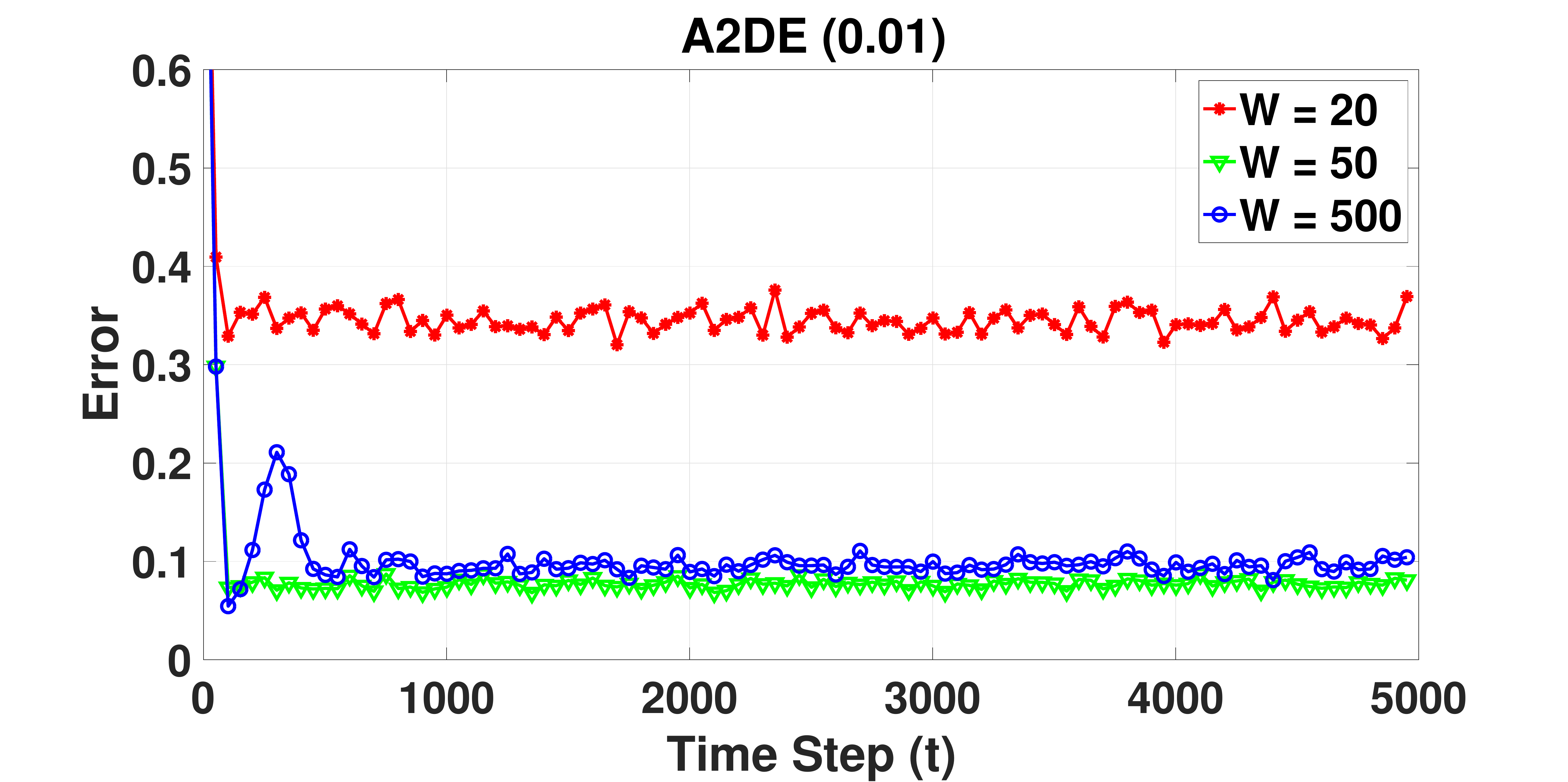}  \caption{} \label{fig_wmd_a2de} \end{subfigure}
\begin{subfigure}[b]{0.49\textwidth}\includegraphics[width=60mm,height=45mm]{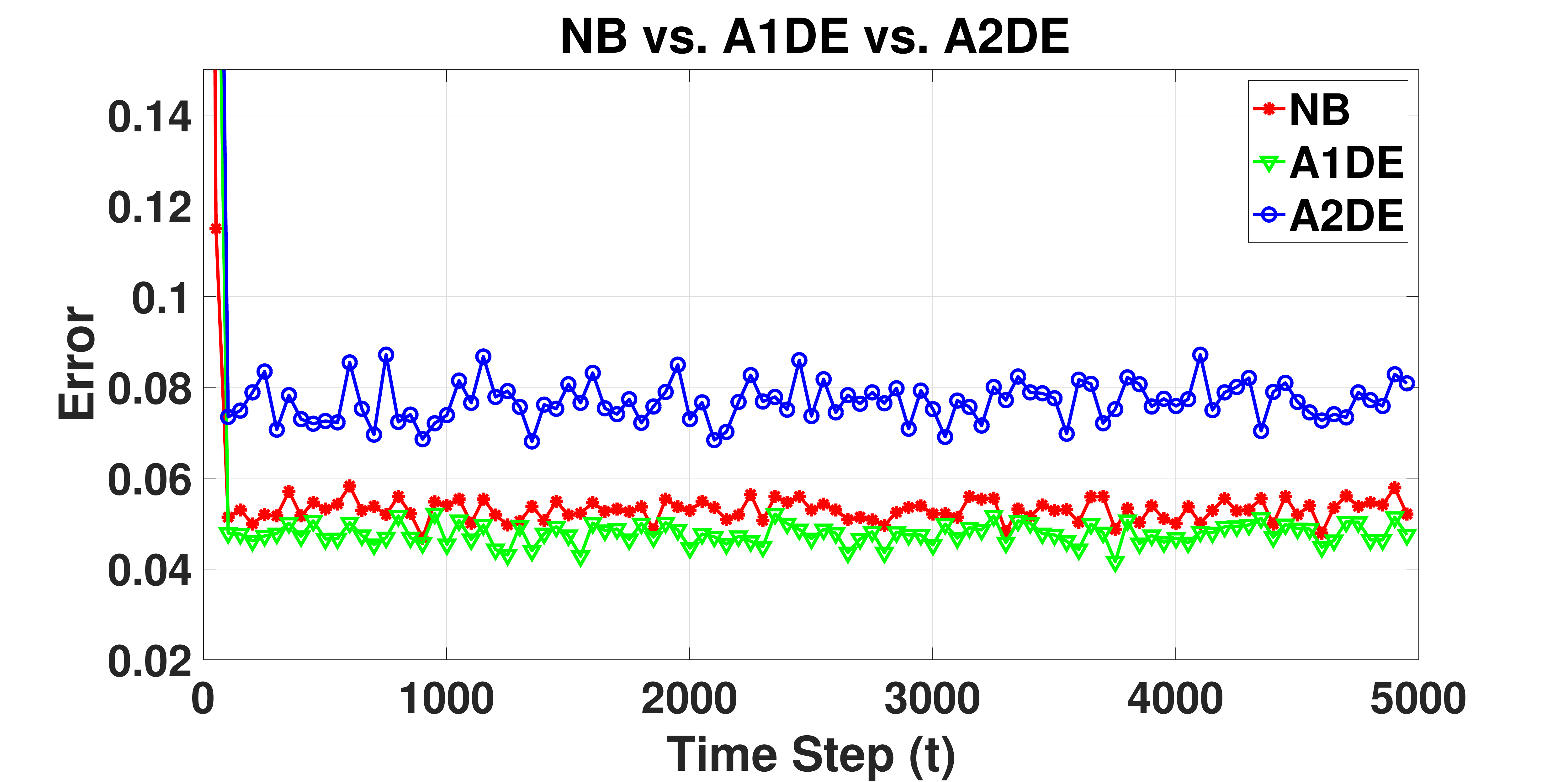} \caption{} \label{fig_wmd_all} \end{subfigure}

\vspace*{-5pt}\caption{\small Windowing with Medium Drift ($\Delta=0.01$) -- Variation in prequential loss of NB (Figure~\ref{fig_wmd_nb}), A1DE (Figure~\ref{fig_wmd_a1de}) and A2DE (Figure~\ref{fig_wmd_a2de}) with  window sizes of 20, 50 and 500. Figure~\ref{fig_wmd_all}: Comparison of NB (error = $0.0531$), A1DE (error = $0.047$) and A2DE (error = $0.083$) with best window size.}

\label{fig_window_mediumDrfit}
\end{figure}
\begin{figure}
\centering

\begin{subfigure}[b]{0.49\textwidth}\includegraphics[width=60mm,height=45mm]{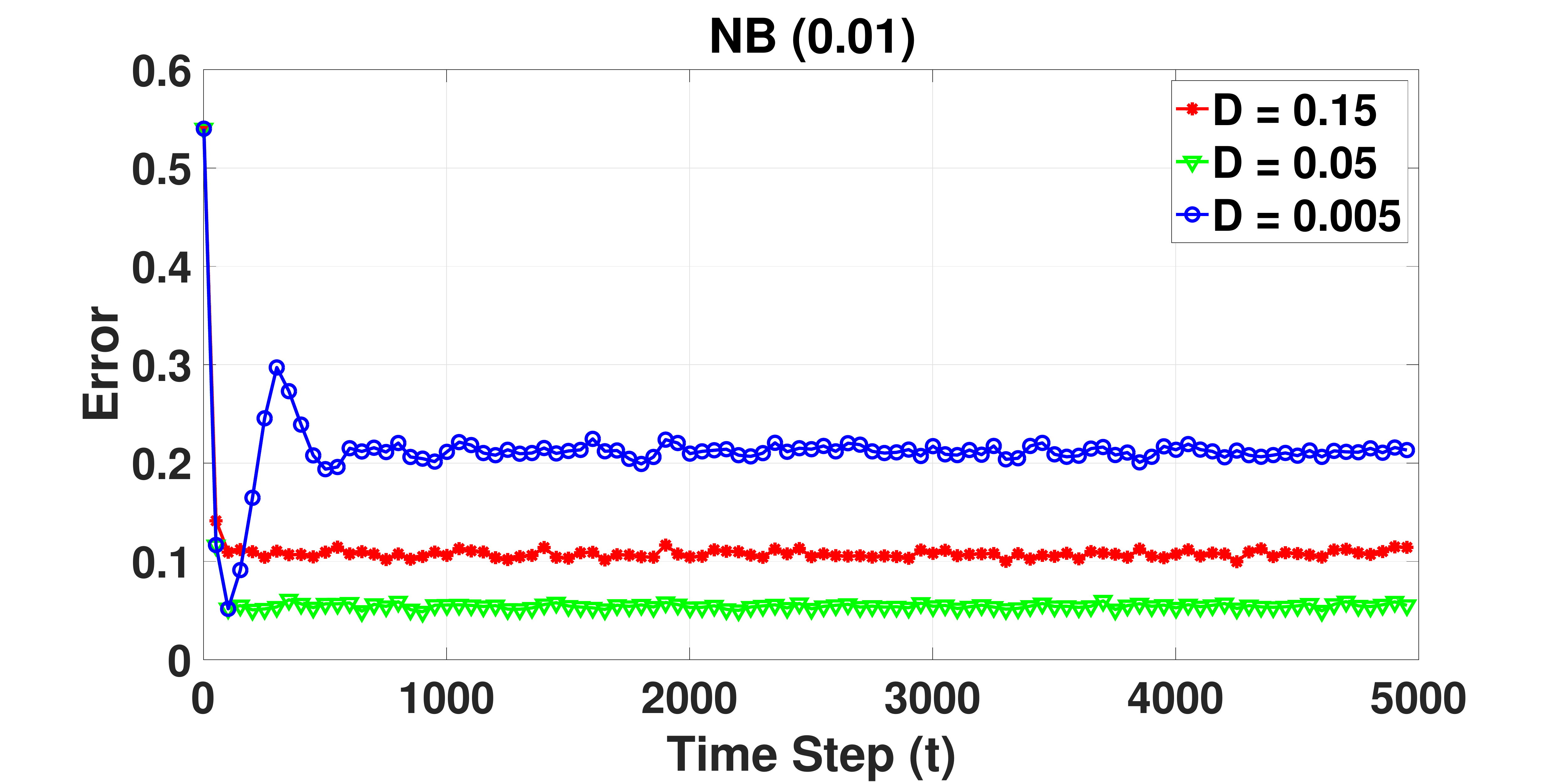} \caption{} \label{fig_dmd_nb} \end{subfigure}
\begin{subfigure}[b]{0.49\textwidth}\includegraphics[width=60mm,height=45mm]{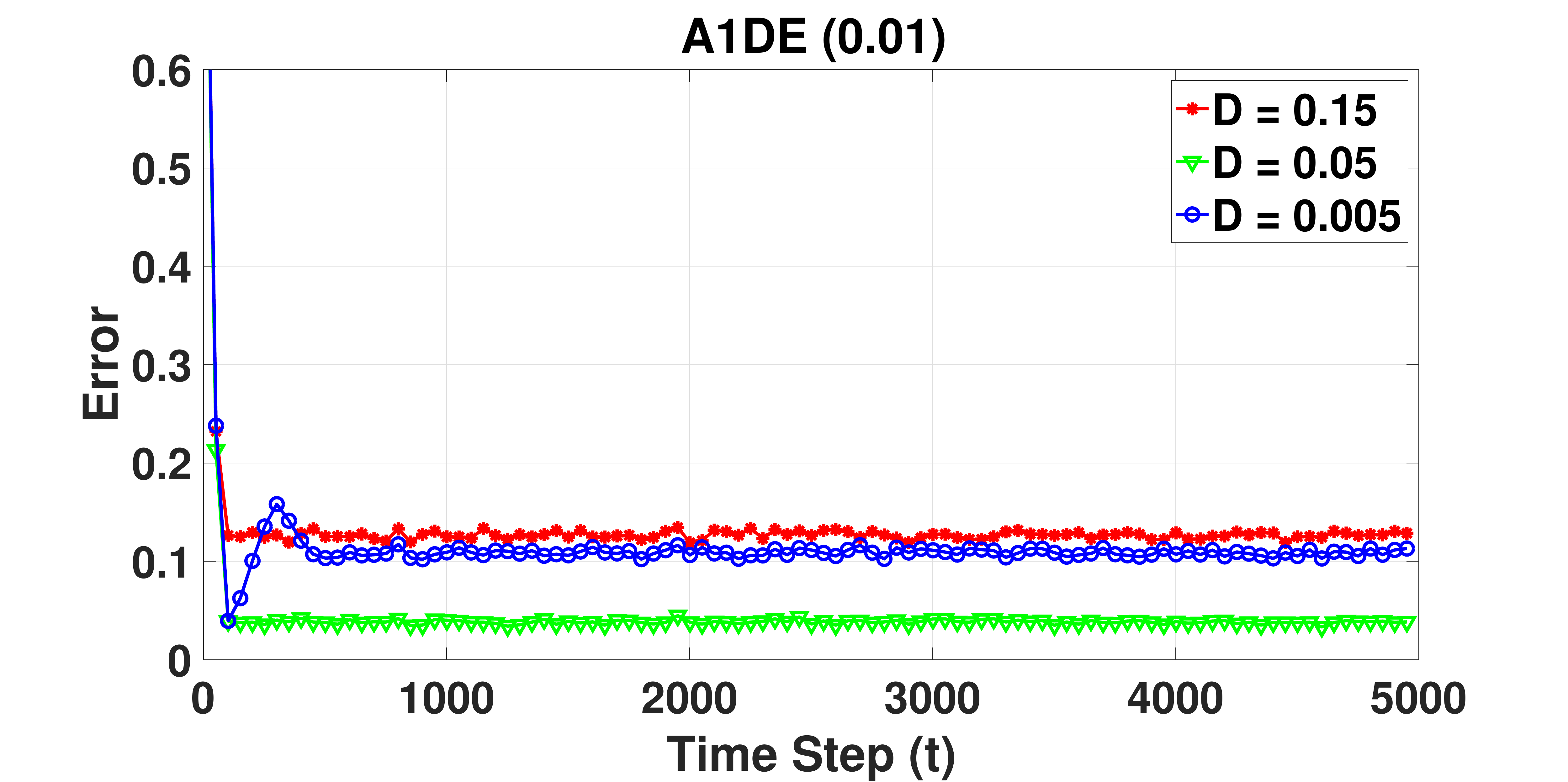} \caption{} \label{fig_dmd_a1de} \end{subfigure}

\begin{subfigure}[b]{0.49\textwidth}\includegraphics[width=60mm,height=45mm]{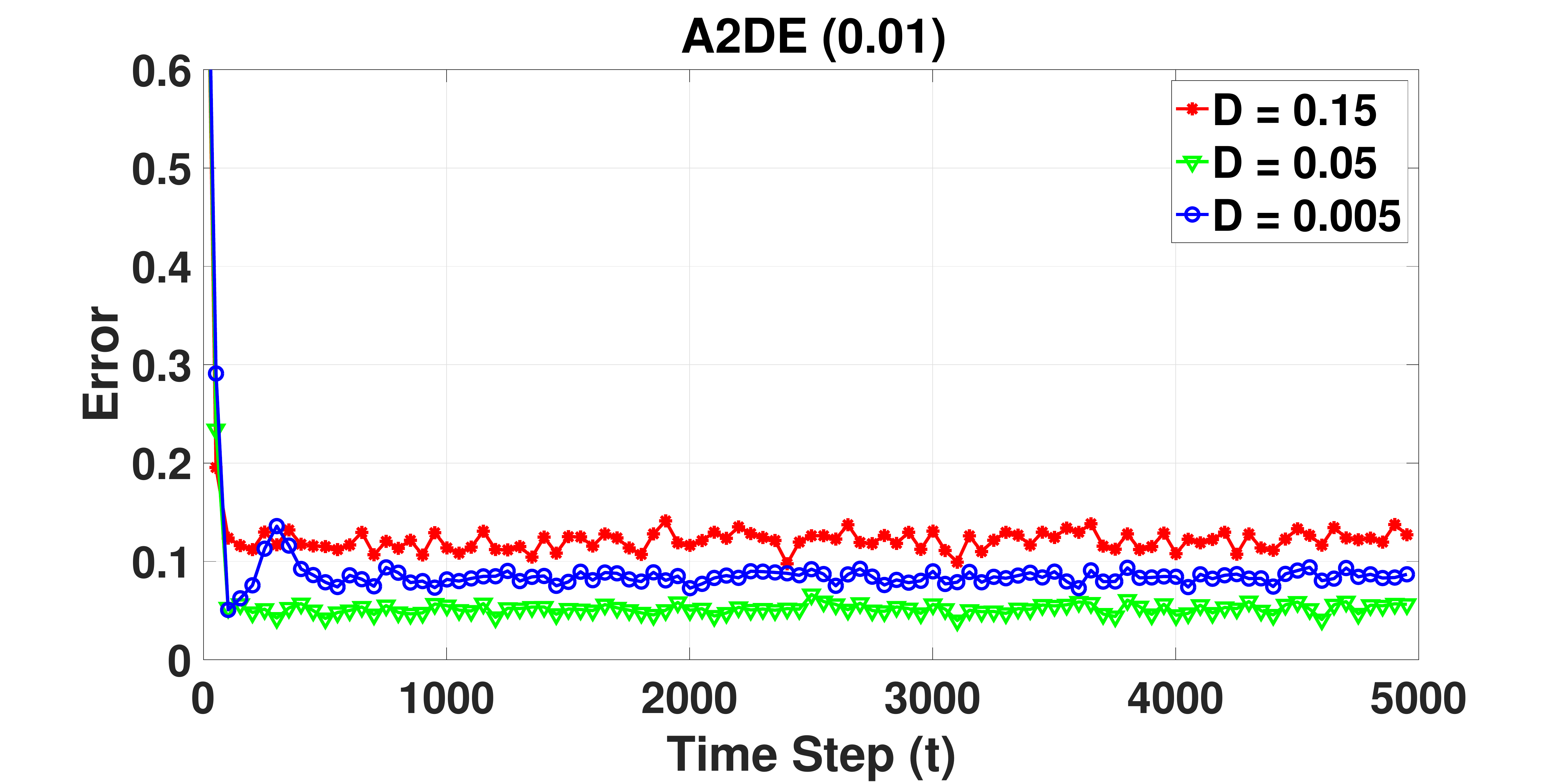} \caption{} \label{fig_dmd_a2de} \end{subfigure}
\begin{subfigure}[b]{0.49\textwidth}\includegraphics[width=60mm,height=45mm]{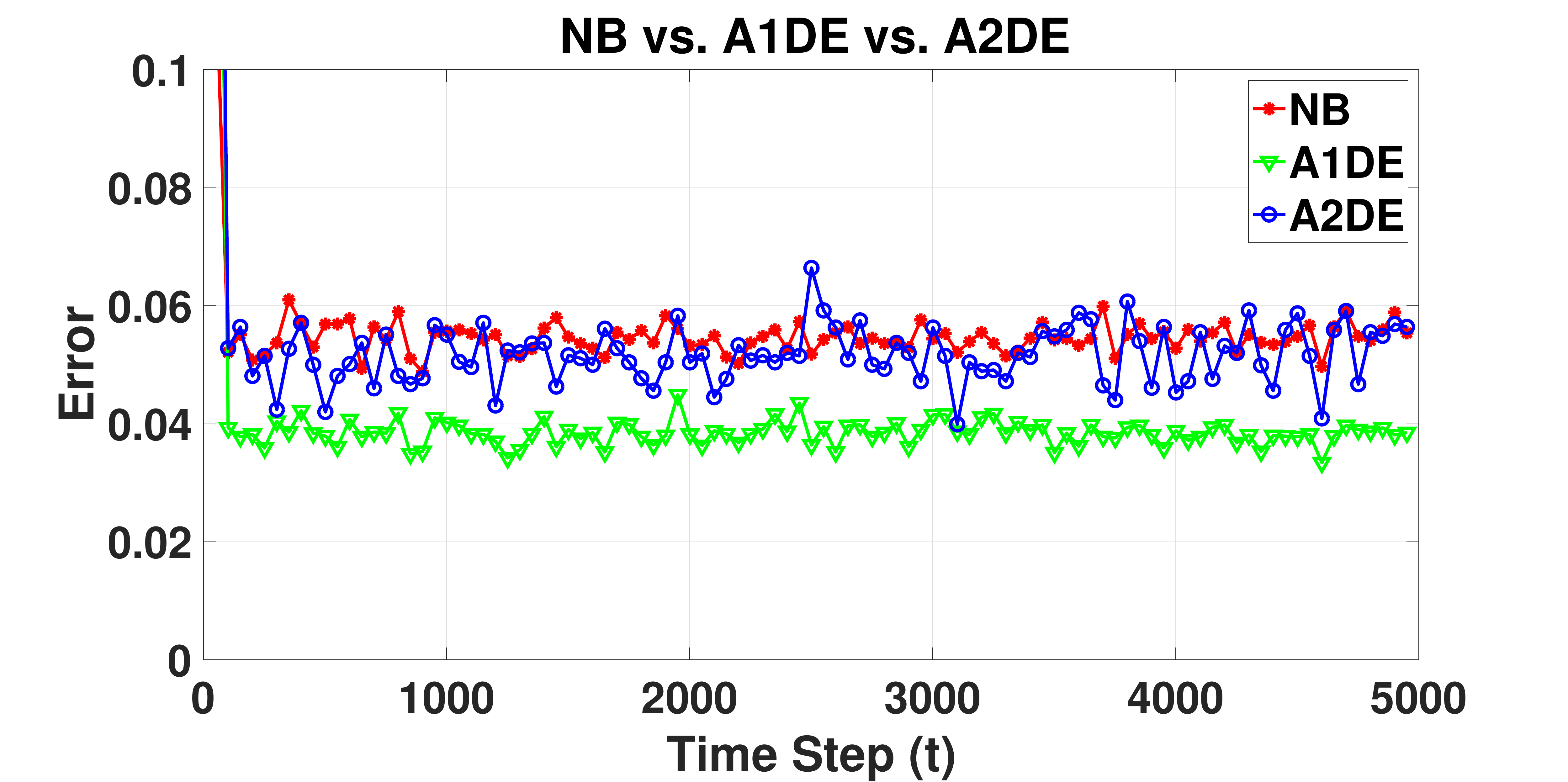} \caption{} \label{fig_dmd_all} \end{subfigure}

\vspace*{-5pt}\caption{\small Decay with Medium Drift ($\Delta=0.01$) -- Variation in prequential loss of NB (Figure~\ref{fig_dmd_nb}), A1DE (Figure~\ref{fig_dmd_a1de}) and A2DE (Figure~\ref{fig_dmd_a2de}) with varying decay rates of $0.15$, $0.05$ and $0.005$. Figure~\ref{fig_dmd_all}: Comparison of NB (error = $0.0599$), A1DE (error = $0.0498$) and A2DE (error = $0.0629$) with best decay rates.}

\label{fig_decay_mediumDrfit}
\end{figure}
At an intermediate drift rate ($\Delta < 0.25$), the intermediate bias learner A1DE starts to outperform NB. Figures~\ref{fig_window_mediumDrfit} and~\ref{fig_decay_mediumDrfit}  show prequential 0-1 Loss  with different window sizes and decay rates with a drift delta of size $0.01$.
It is apparent that for this intermediate drift rate the intermediate window size (50) and intermediate decay rate (0.05) attain the lowest error and that the learner with intermediate bias (A1DE) minimizes overall error. Figures~\ref{fig_wmd_all} and~\ref{fig_dmd_all}, shows the comparison of NB, A1DE and A2DE with their best respective window size and decay rate, and it can be seen that A1DE results in best performance.

\begin{figure}
\centering

\begin{subfigure}[b]{0.49\textwidth}\includegraphics[width=60mm,height=45mm]{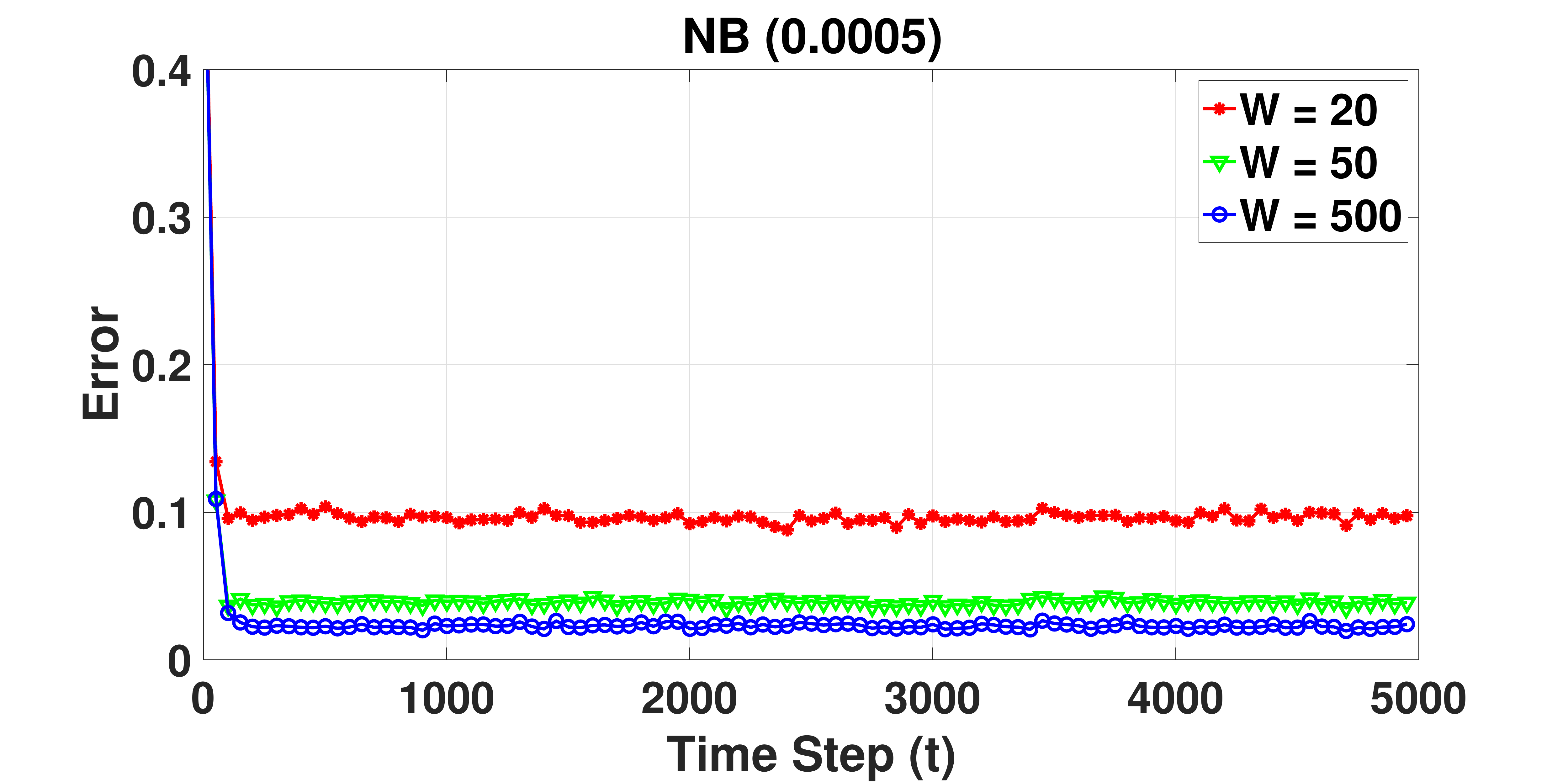} \caption{} \label{fig_wsd_nb} \end{subfigure}
\begin{subfigure}[b]{0.49\textwidth}\includegraphics[width=60mm,height=45mm]{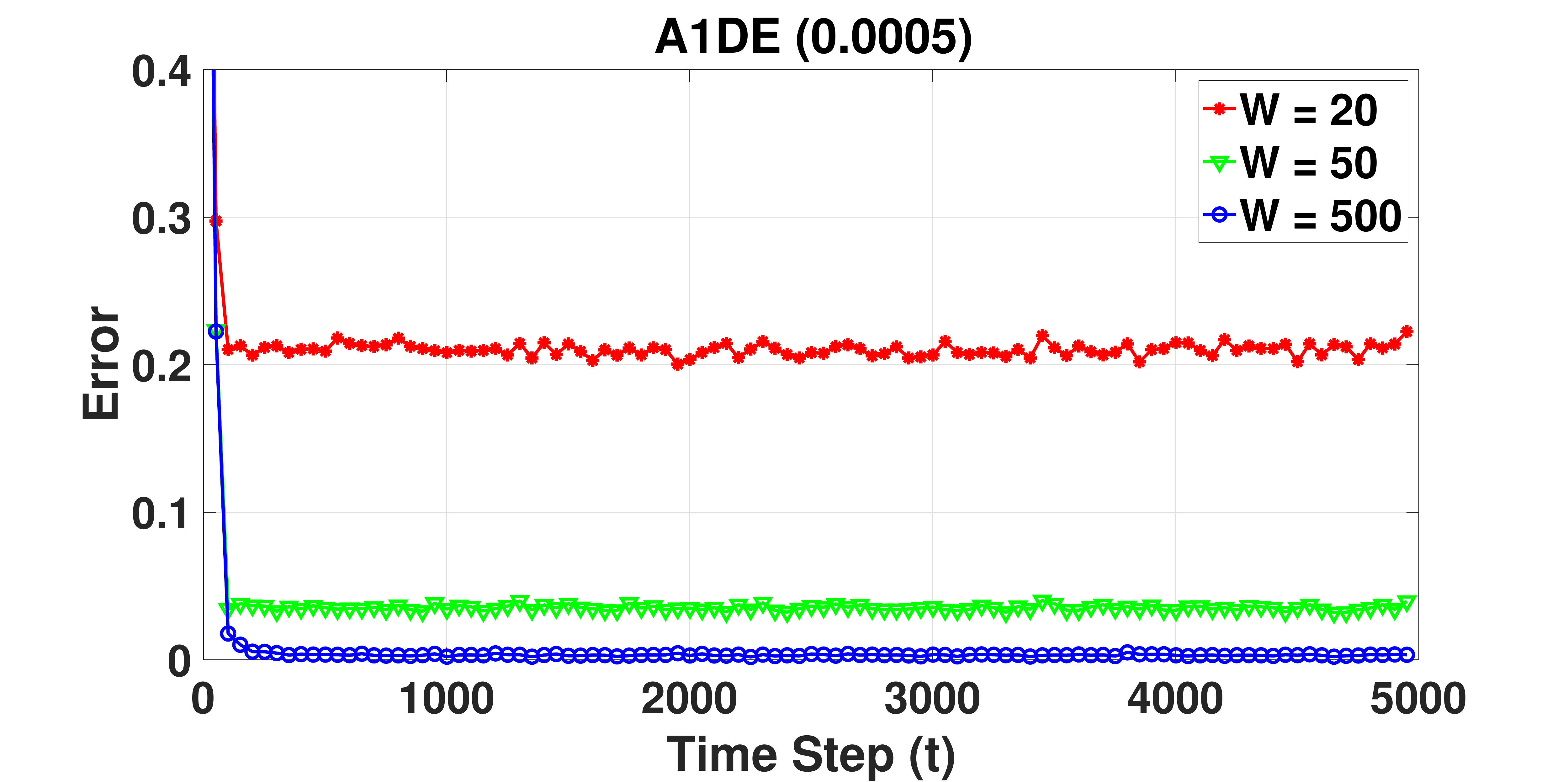} \caption{} \label{fig_wsd_a1de} \end{subfigure}

\begin{subfigure}[b]{0.49\textwidth}\includegraphics[width=60mm,height=45mm]{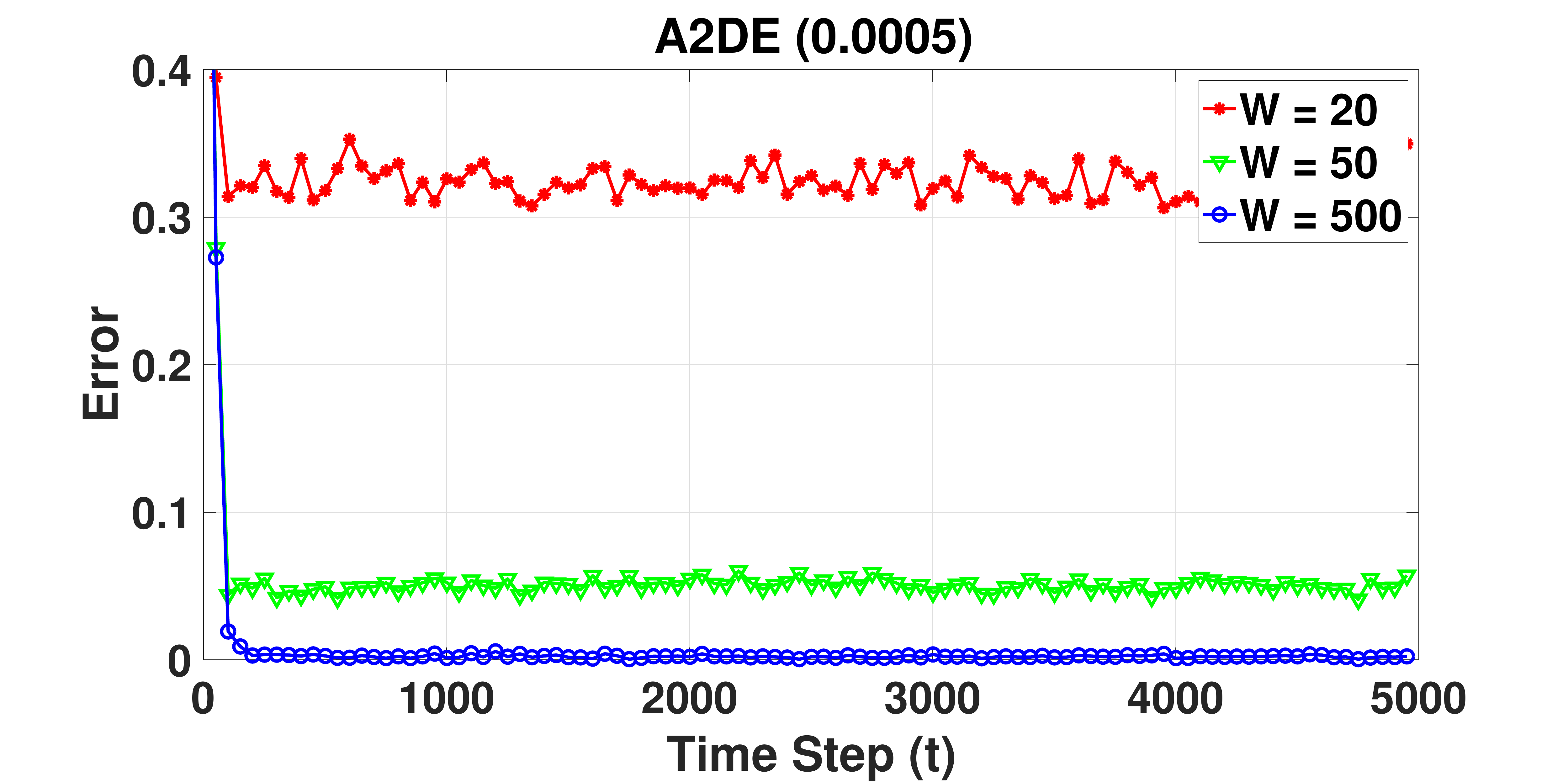} \caption{} \label{fig_wsd_a2de} \end{subfigure}
\begin{subfigure}[b]{0.49\textwidth}\includegraphics[width=60mm,height=45mm]{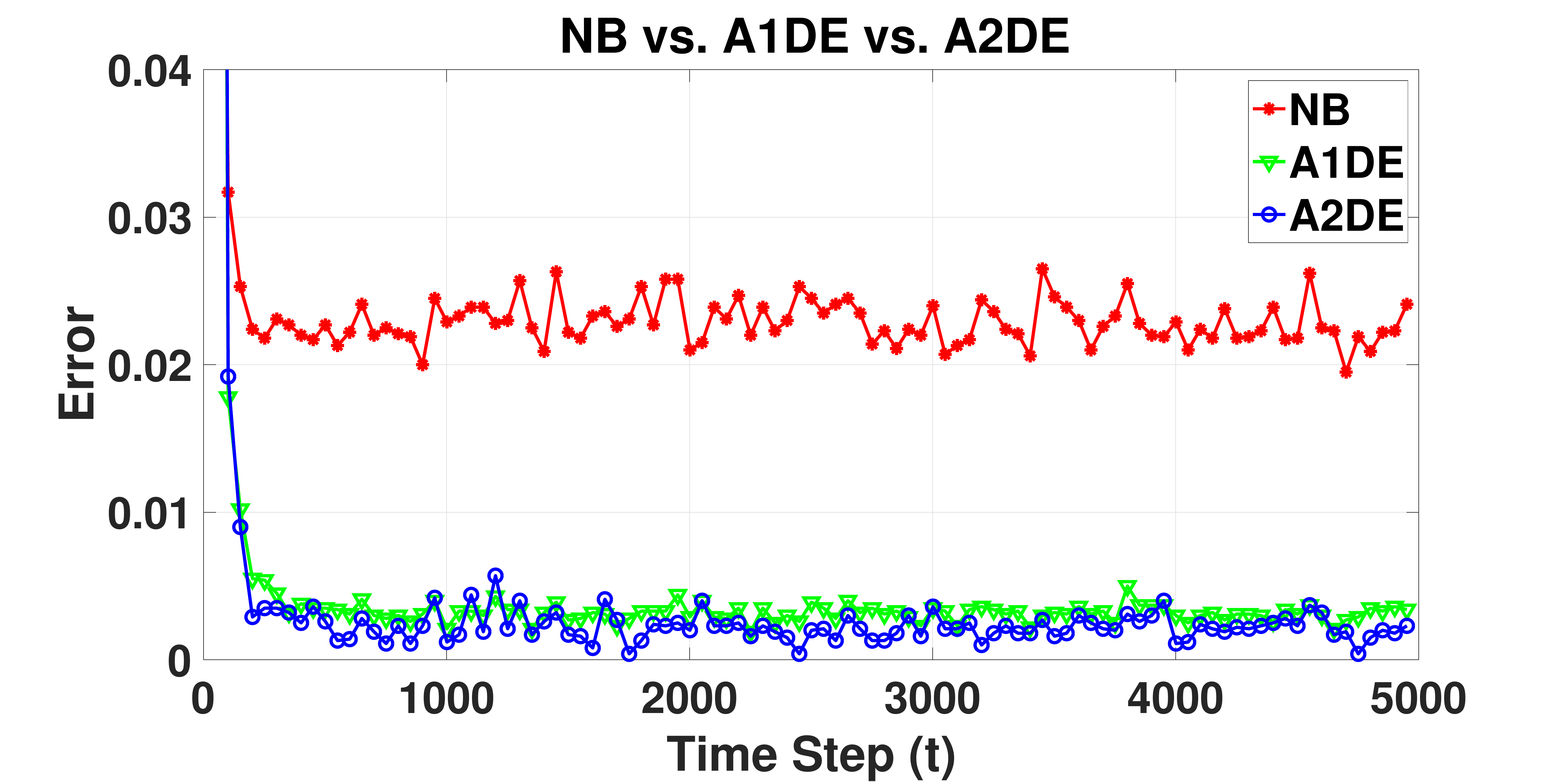} \caption{} \label{fig_wsd_all} \end{subfigure}

\vspace*{-5pt}\caption{\small Windowing with Slow Drift ($\Delta=0.0005$) -- Variation in prequential loss of NB (Figure~\ref{fig_wsd_nb}), A1DE (Figure~\ref{fig_wsd_a1de}) and A2DE (Figure~\ref{fig_wsd_a2de}) with window sizes of 20, 50 and 500. Figure~\ref{fig_wsd_all}: Comparison of NB  (error = $0.0289$), A1DE (error = $0.0156$) and A2DE (error = $0.0151$) with best window size.}

\label{fig_window_slowDrfit}
\end{figure}
\begin{figure}
\centering

\begin{subfigure}[b]{0.49\textwidth}\includegraphics[width=60mm,height=45mm]{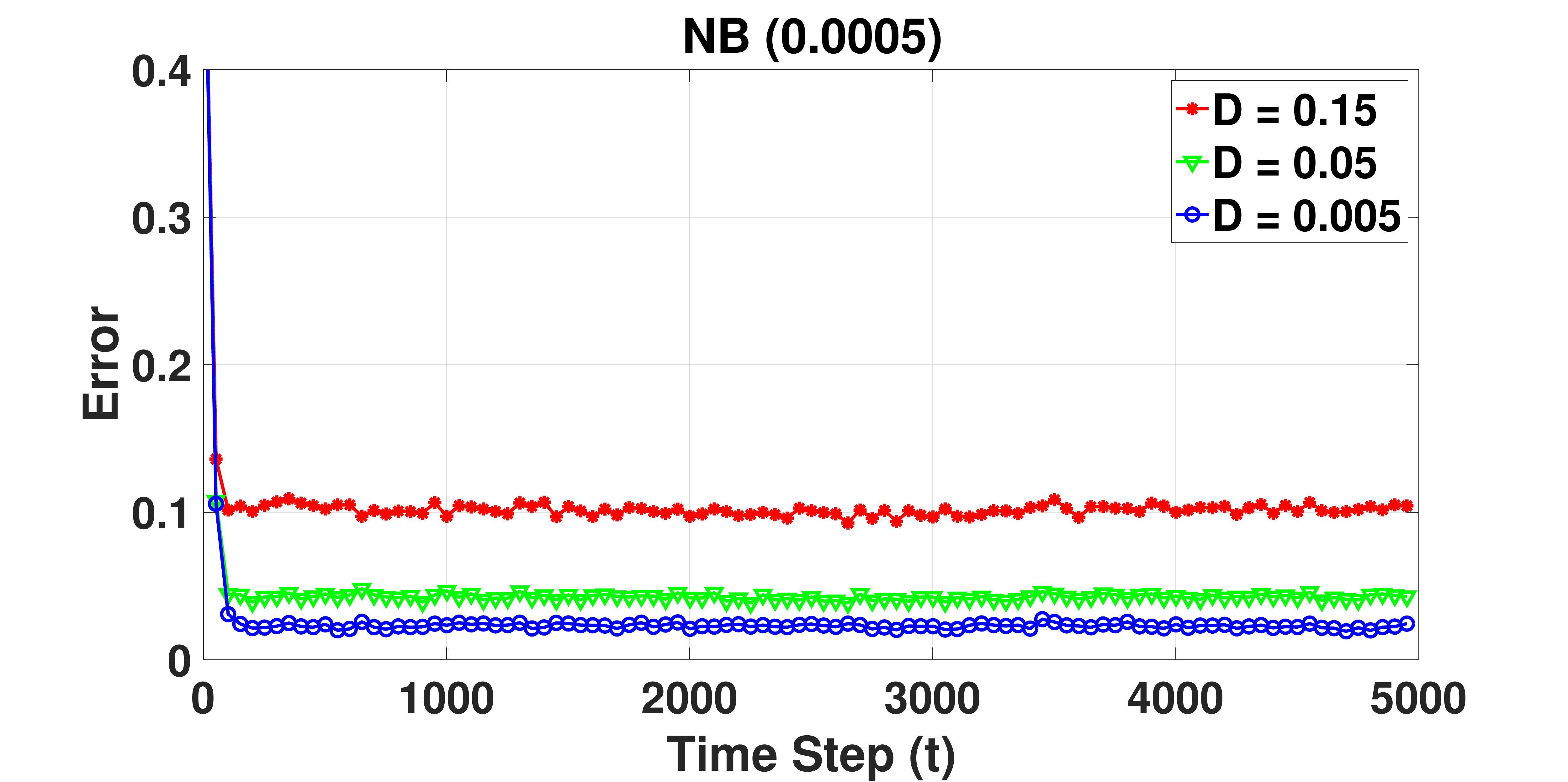} \caption{} \label{fig_dsd_nb} \end{subfigure}
\begin{subfigure}[b]{0.49\textwidth}\includegraphics[width=60mm,height=45mm]{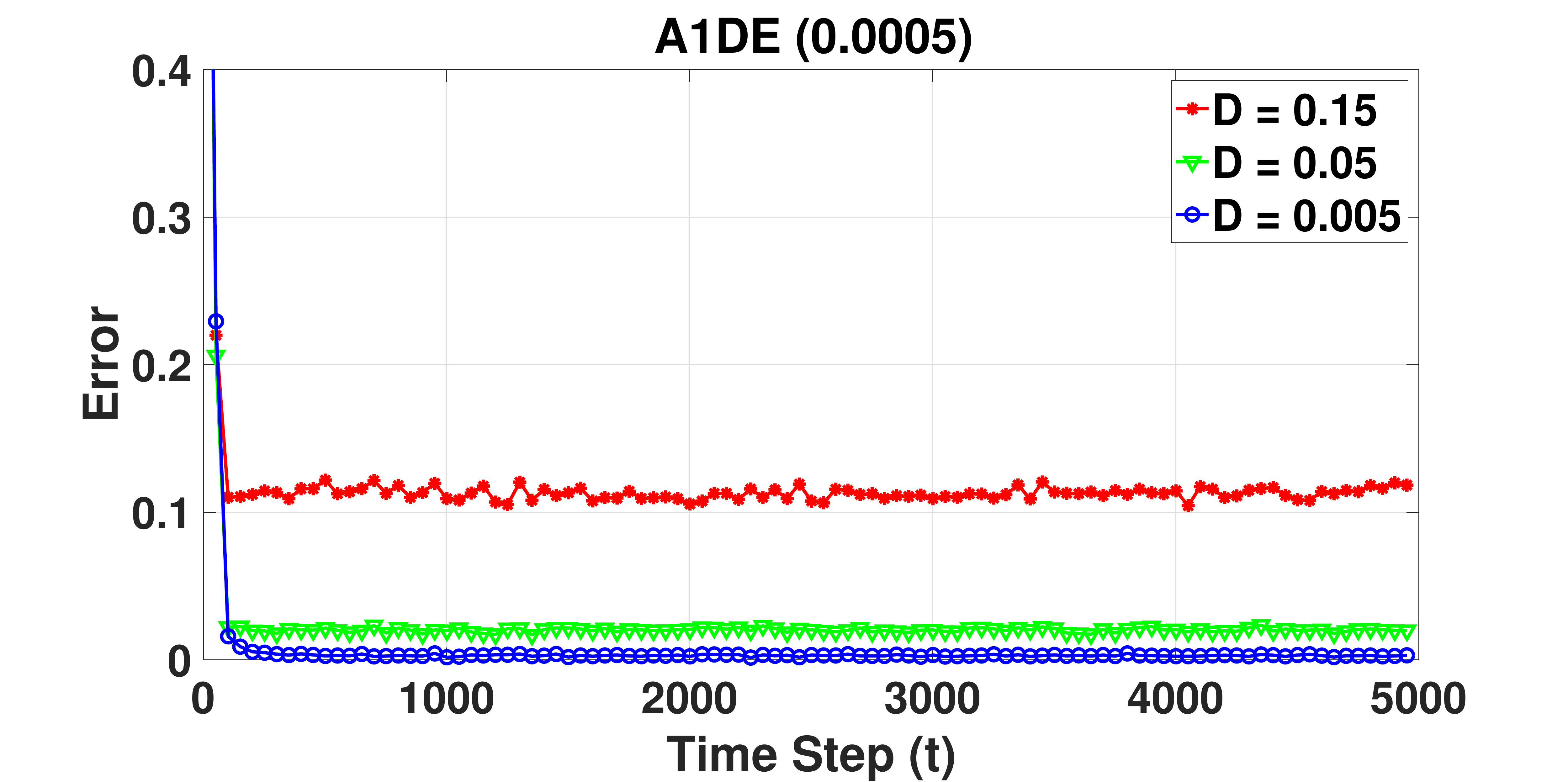} \caption{} \label{fig_dsd_a1de} \end{subfigure}

\begin{subfigure}[b]{0.49\textwidth}\includegraphics[width=60mm,height=45mm]{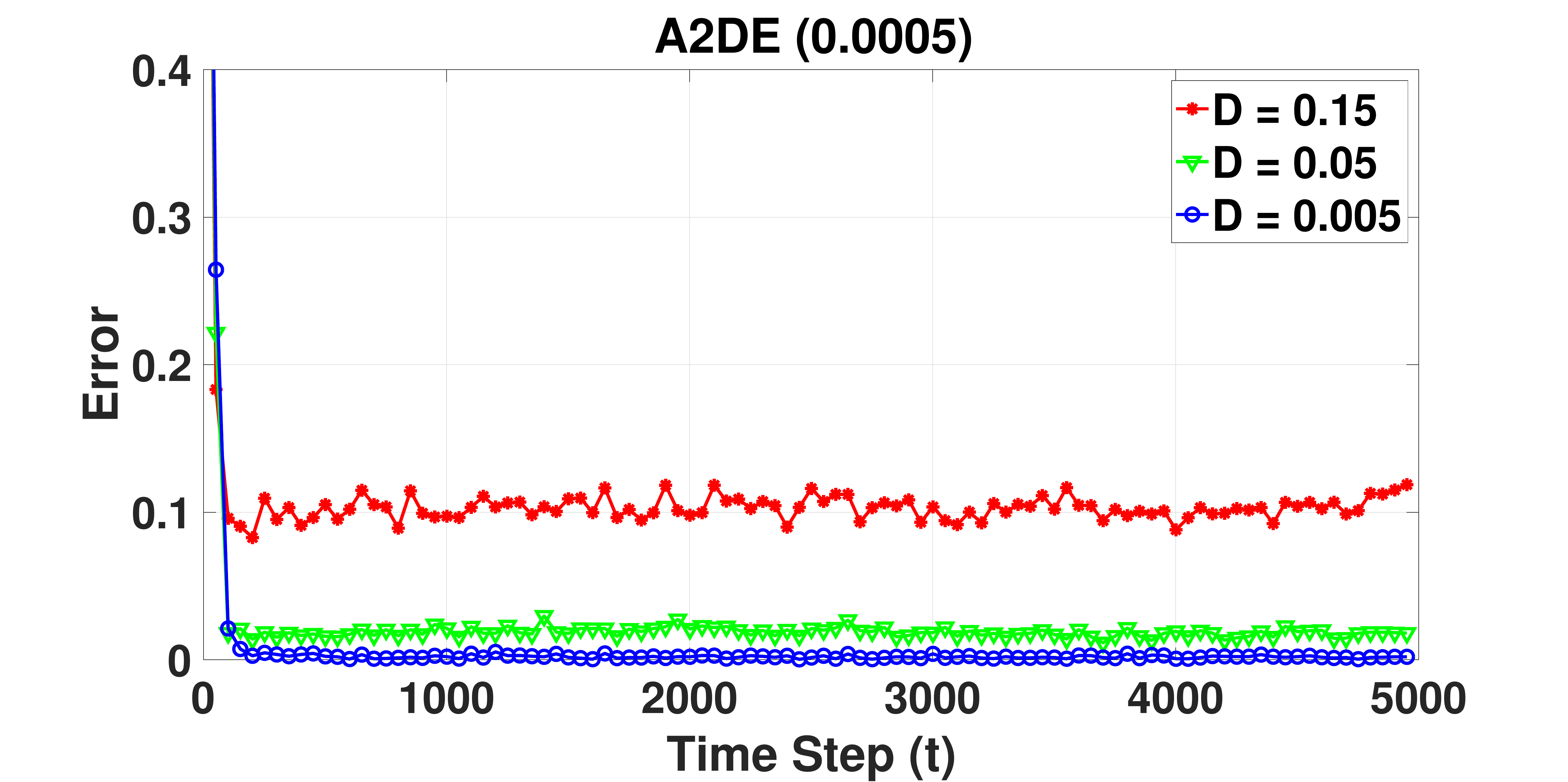} \caption{} \label{fig_dsd_a2de} \end{subfigure}
\begin{subfigure}[b]{0.49\textwidth}\includegraphics[width=60mm,height=45mm]{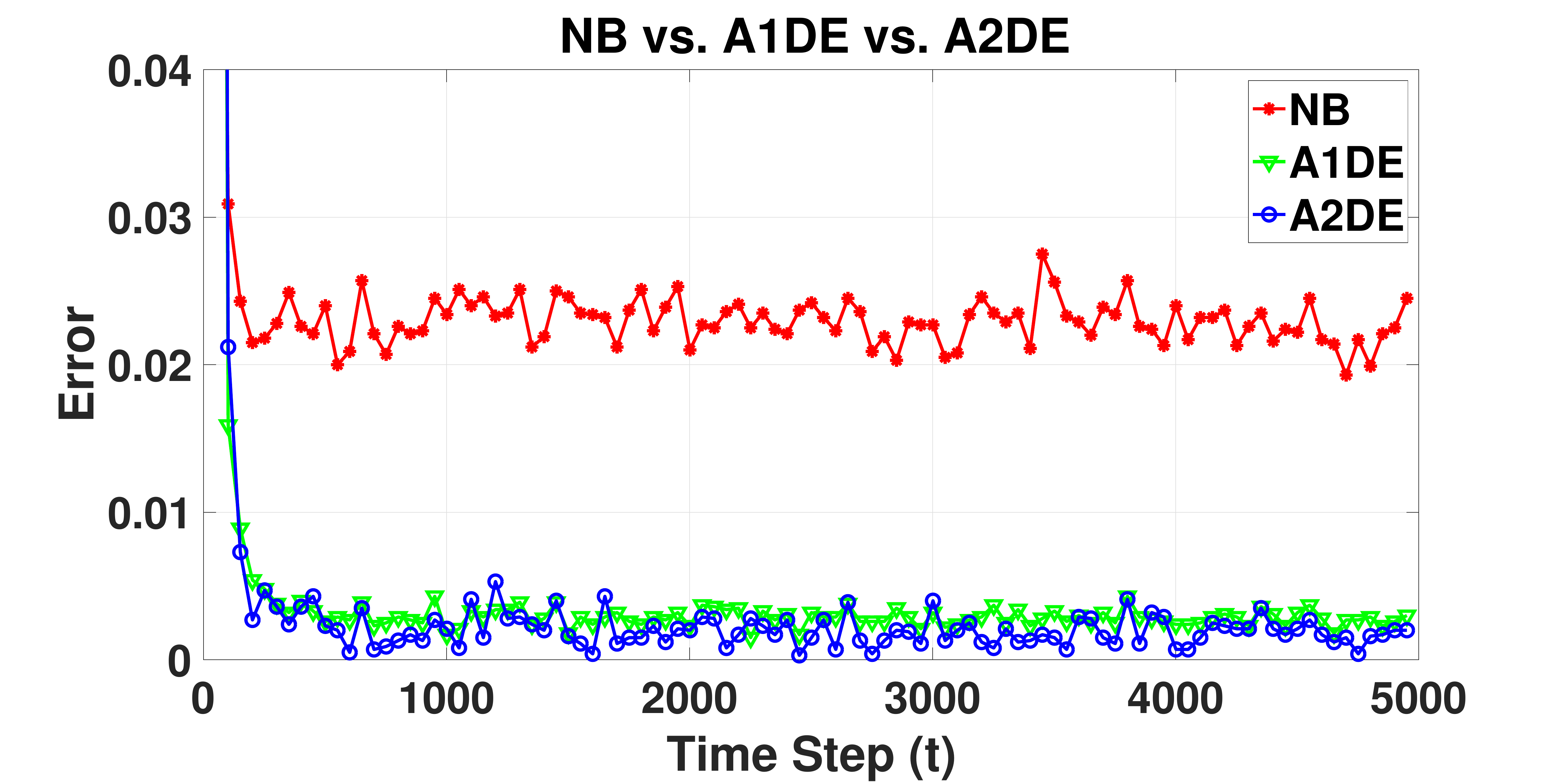} \caption{} \label{fig_dsd_all} \end{subfigure}

\vspace*{-5pt}\caption{\small Decay with Slow Drift ($\Delta=0.0005$) -- Variation in 0-1 Loss of NB (Figure~\ref{fig_dsd_nb}), A1DE (Figure~\ref{fig_dsd_a1de}) and A2DE (Figure~\ref{fig_dsd_a2de}) with varying decay rates of $0.15$, $0.05$ and $0.005$. Figure~\ref{fig_dsd_all}: Comparison of NB (error = $0.0290$), A1DE (error = $0.0153$) and A2DE (error = $0.0148$) with best decay size.}

\label{fig_decay_slowDrfit}
\end{figure}
Figures~\ref{fig_window_slowDrfit} and~\ref{fig_decay_slowDrfit} show prequential error with a slow drift rate, $\Delta=0.0005$, with varying window sizes and decay rates. Figure~\ref{fig_wsd_all}, compares the performance of the three models with their respective best window size and Figure~\ref{fig_dsd_all} with their best decay rate. In both cases it is apparent that A2DE achieves the lowest error. 

Thus, in all three scenarios of fast, intermediate and slow drift we find the relationship predicted by the sweet path between drift rate, forgetting rate and bias/variance profile.


\subsection{Exploiting the insights of the sweet path for designing practical learners} \label{subsec_soa}

We have demonstrated that a generalizable and falsifiable hypothesis is consistent with experimental outcomes. This raises the question of how the resulting insights might be used to create new effective learners that can respond to concept drift. Unfortunately, doing so appears to be far from trivial. If we allow that drift rates may vary over time, the sweet path suggests that we need learners that can adapt to changes in drift rates with corresponding adaptation in their forgetting rates and bias/variance profiles.  We are yet to devise a practical algorithmic solution to the complex problem.

\clearpage
To assess the practical implications of our hypotheses for designing practical learning algorithms, we compare the performance on real-world drifting data of our AnDE classifiers with differing forgetting rates and bias/variance profiles against  a range of  state of the art concept drift classifiers. We use $4$ standard benchmark drift datasets, whose details are given in Table~\ref{tab_datasets}. Numeric attributes are discretized using IDAW discretization \citep{Webb14} with 5 intervals.
\begin{table} \scriptsize
\center
\tabcolsep=2.0pt \renewcommand{\arraystretch}{1.3}
\begin{tabular}{lcccccc}
\cline{1-4}
		& \#Instances 	& \#Attributes  & \#Classes \\
\cline{1-4}
PowerSupply 	& 29928		& 2 	& 2 \\
Airlines 		& 539383	& 7 	& 2 \\
ElectricNorm 	& 45312		& 8 	& 2 \\
Sensor 			& 2219803	& 5 	& 58\\
\cline{1-4}
\end{tabular}
\vspace{-0.05in}
\caption{\small Details of datasets.}
\label{tab_datasets}
\end{table}

We compare the performance with $12$ standard learning techniques:
\begin{enumerate}
\item AccuracyUpdatedEnsemble~\citep{AccuracyUpdatedEnsemble}, 
\item AccuracyWeightedEnsemble~\citep{AccuracyWeightedEnsemble}, 
\item DriftDetectionMethodClassifier~\citep{DDM}, 
\item DriftDetectionMethodClassifierEDDM~\citep{EDDM}, 
\item HoeffdingAdaptiveTree~\citep{HoeffdingAdaptiveTree},
\item HoeffdingOptionTree~\citep{HoeffdingOptionTree},
\item HoeffdingTree~\citep{HoeffdingTree}, 
\item LeveragingBag~\citep{LeveragingBag},  
\item OzaBag~\citep{OzaBagBoost}, 
\item OzaBagAdwin~\citep{OzaBagAdwin},
\item OzaBoost~\citep{OzaBagBoost},
\item OzaBoostAdwin~\citep{OzaBagBoost,ADWIN}.
\end{enumerate}
It can be seen from Table~\ref{tab_TM_Results} that  
\begin{table} \scriptsize
\center
\tabcolsep=1.0pt \renewcommand{\arraystretch}{1.3}
\begin{tabular}{lccccc}
\cline{1-5}
& AccUpdatedEns & OzaBagAdwin	& DriftDetClassifier & DriftDetClassifierEDDM \\
\cline{2-5}
PowerSupply & 0.8599 & 0.8692 & 0.8634 & 0.8615 \\
Airlines & \bf 0.3335 & 0.3448 & 0.3534 & 0.3511 \\ 
ElectricNorm & 0.2219 & 0.167 & 0.1984 & 0.149 \\ 
Sensor & 0.3102 & 0.2874 & 0.3206 & 0.3159 \\ 
\cline{2-5}
& ASHoeffdingTree & HoeffdingTree & OzaBag & HoeffdingAdaptiveTree \\
\cline{2-5}
PowerSupply & 0.864 & 0.864 & 0.8655 & 0.8661 \\
Airlines & 0.3552 & 0.3552 & 0.3575 & 0.3632 \\ 
ElectricNorm & 0.2007 & 0.2007 & 0.1982 & 0.1759 \\ 
Sensor & 0.7153 & 0.7153 & 0.7067 & 0.3718 \\
\cline{2-5}
& OzaBoost	& AccWeightedEns	& LeveragingBag & OzaBoostAdwin	\\
\cline{2-5}
PowerSupply & 0.9583 & \bf 0.8579 & 0.8717 & 0.9584 \\
Airlines & 0.3719 & 0.3751 & 0.3769 & 0.3888 \\
ElectricNorm & 0.1781 & 0.2471 & \bf 0.1303 & 0.143 \\
Sensor & 0.9514 & 0.3596 & \bf 0.2395 & 0.407 \\
\cline{1-5}
\cline{1-5}
\end{tabular}
\vspace{-0.05in}
\caption{\small Comparison of 0-1 Loss performance of $12$ standard concept drift techniques on four real-world datasets: \texttt{PowerSupply, Airlines, ElectricNorm, Sensor}. The best results are highlighted in bold font.}
\label{tab_TM_Results}
\end{table}%
\texttt{AccWeightedEns} (AccuracyWeightedEnsemble) achieved the lowest error on~\texttt{PowerSupply}, \texttt{AccUpdatedEn} (AccuracyUpdateEnsemble) the lowest on~\texttt{Airlines}, whereas \texttt{LeveregingBag} (LeveregingBagging) achieved the lowest error on~\texttt{ElectricNorm} and~\texttt{Sensor}.
In the following, we will use these (best) results as benchmarks and see how adaptive AnDE with decay and window based adaptation can perform relative to these results.

\begin{figure}
\centering
\hspace{-0.2in}
\includegraphics[width=60mm,height=45mm]{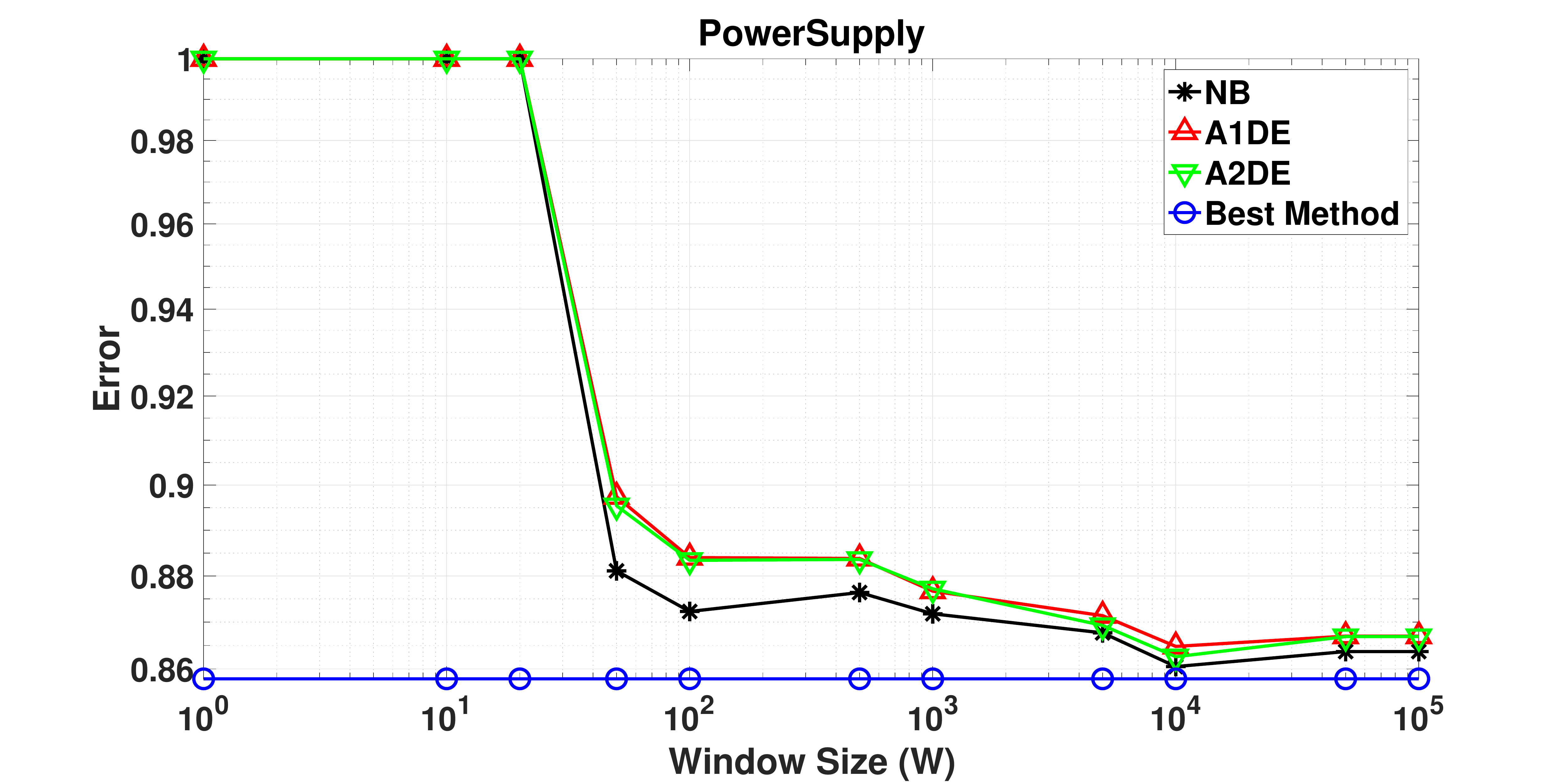}
\includegraphics[width=60mm,height=45mm]{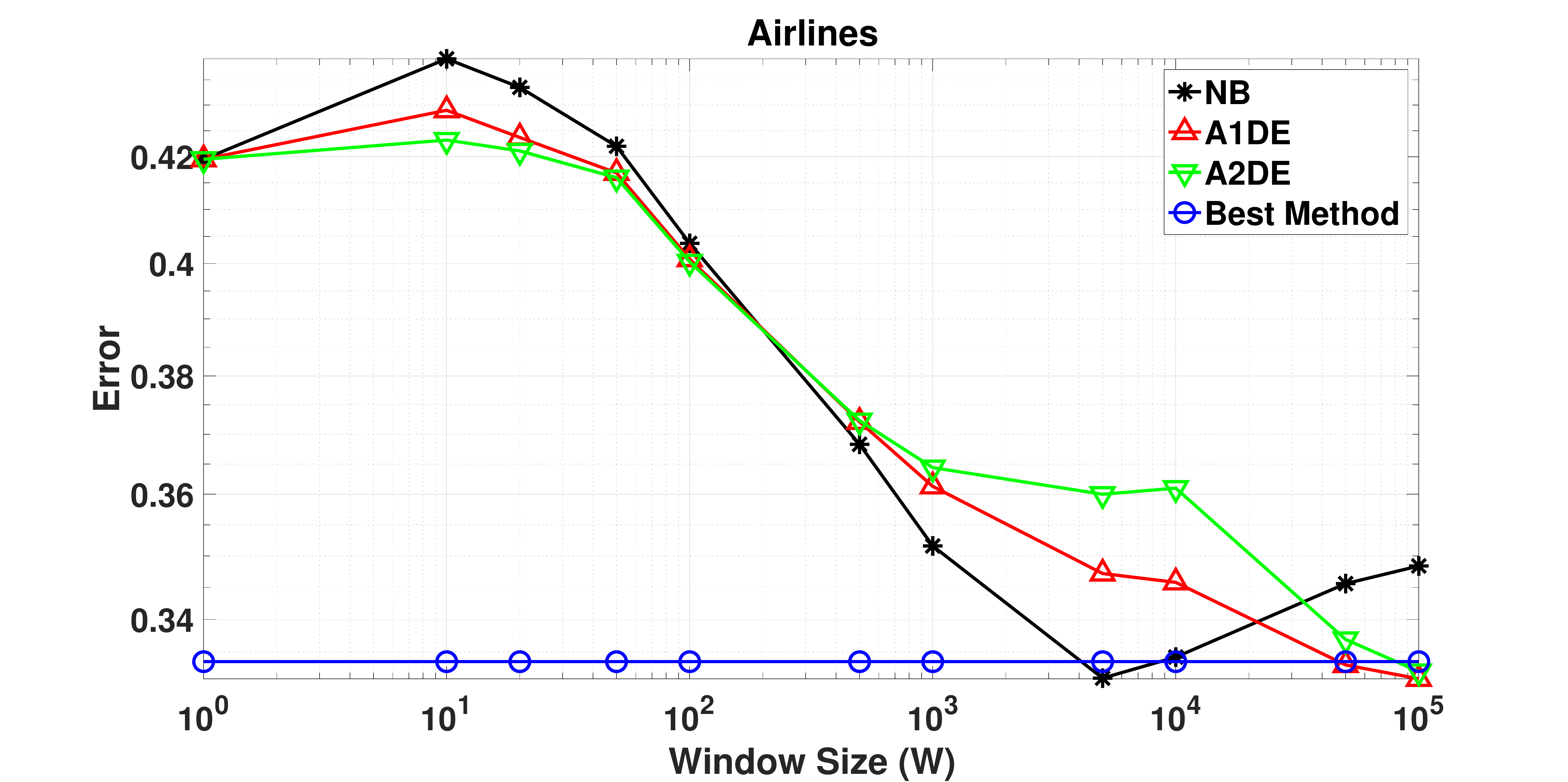}

\hspace{-0.2in}
\includegraphics[width=60mm,height=45mm]{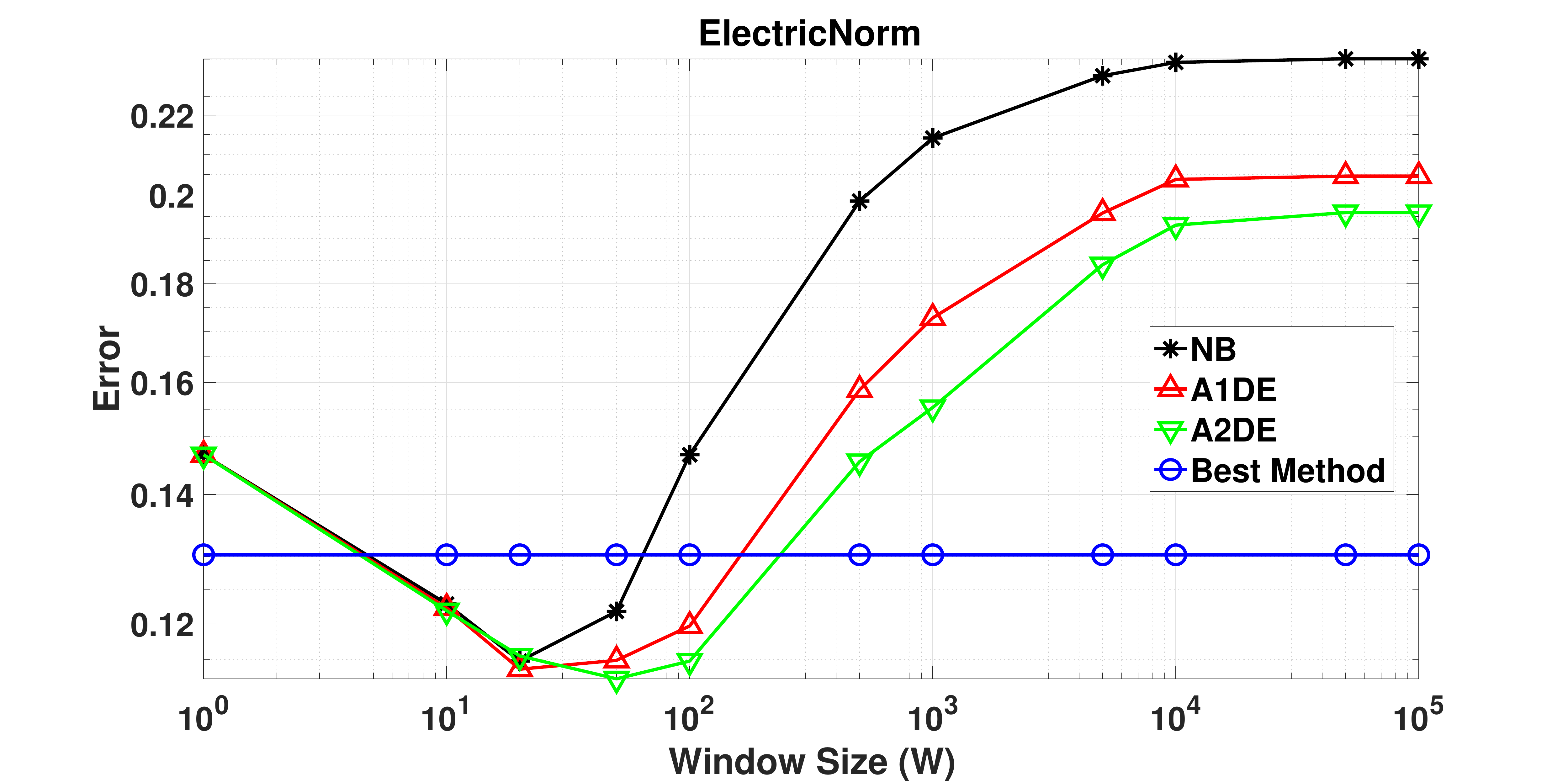}
\includegraphics[width=60mm,height=45mm]{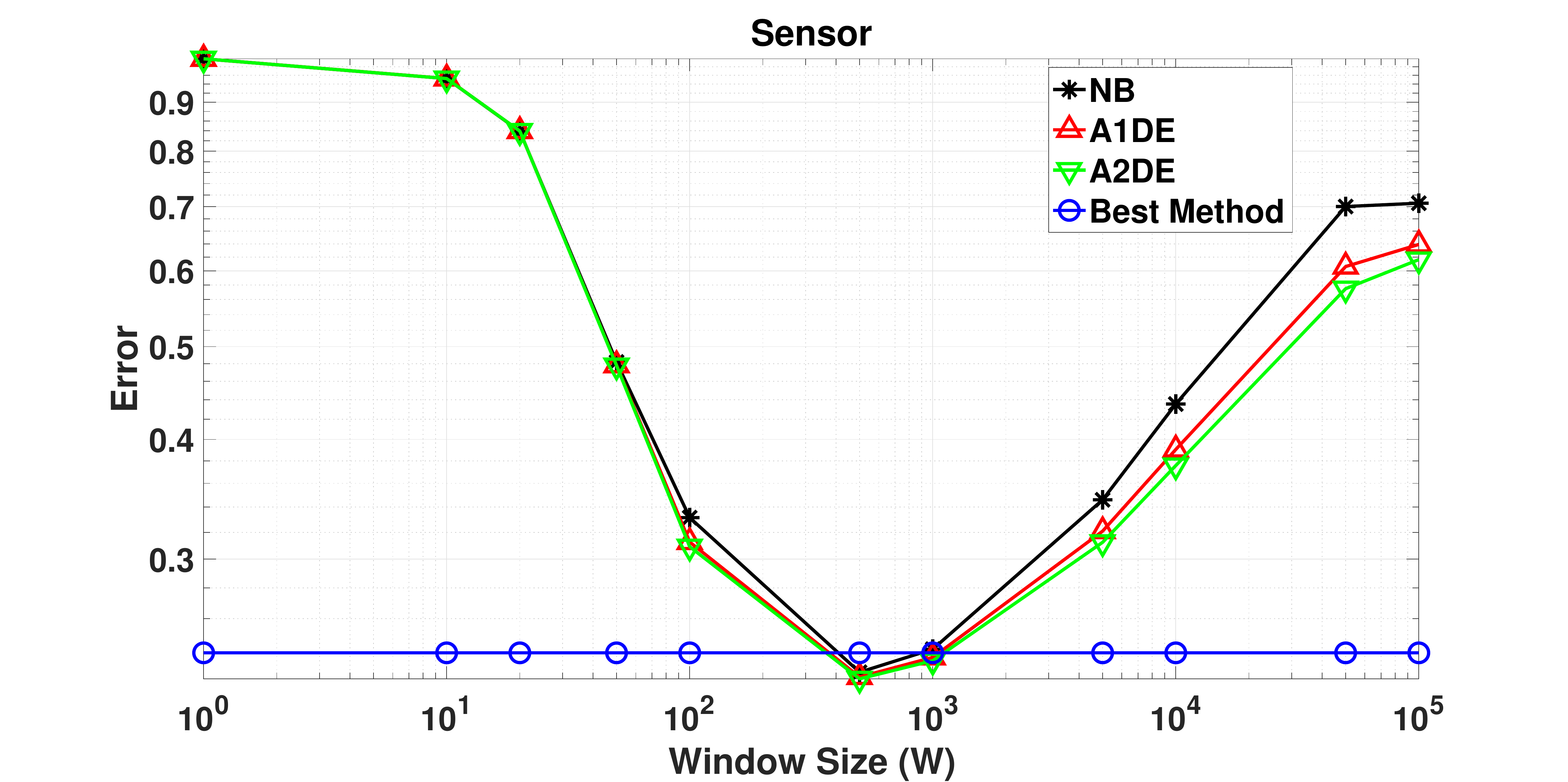}

\vspace*{-5pt}\caption{\small Comparison of adaptive NB, A1DE and A2DE based on window-based adaptation on four real world datasets: \texttt{PowerSupply, Airlines, ElectricNorm, Sensor}. Horizontal blue line depicts the best performance out of $12$ standard techniques.}
\label{fig_RW_Comp_W}
\end{figure}

We compare the performance of adaptive AnDE with window-based adaptation in Figure~\ref{fig_RW_Comp_W}, depicting results with various window sizes. 
The lowest error obtained on the dataset by one of $12$ standard techniques, is also plotted as a horizontal blue line for comparison.
It can be seen that except for~\texttt{PowerSupply}, adaptive AnDE can achieve lower error than the lowest of any of the twelve state-of-the-art techniques. For~\texttt{Airlines}, A1DE achieves the lowest error of $0.3309$ and for~\texttt{ElectricNorm} and~\texttt{Sensors}, A2DE achieves errors of:~$0.1124$ and $0.2250$.

\begin{figure}
\centering
\hspace{-0.2in}
\includegraphics[width=60mm,height=45mm]{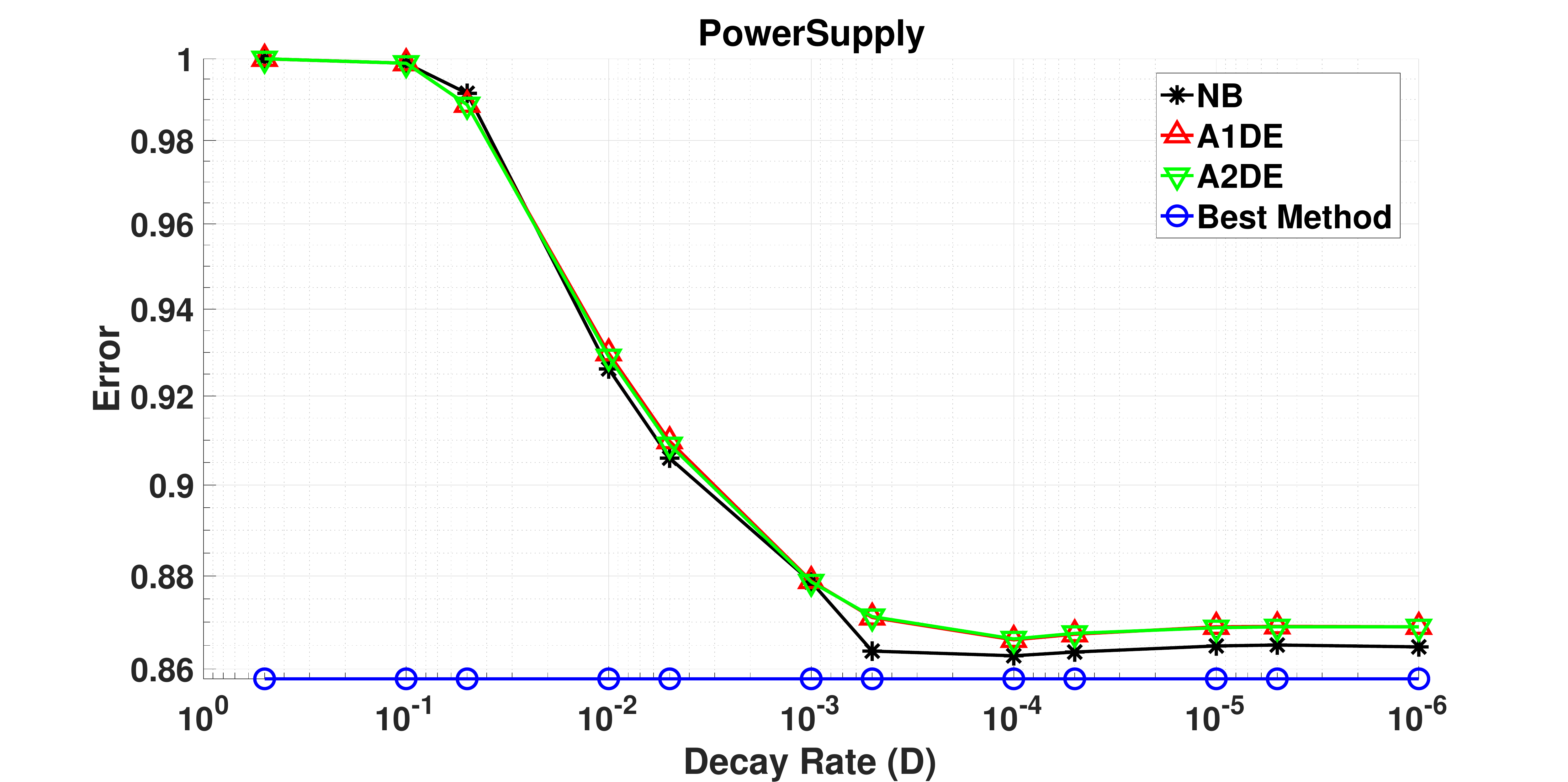}
\includegraphics[width=60mm,height=45mm]{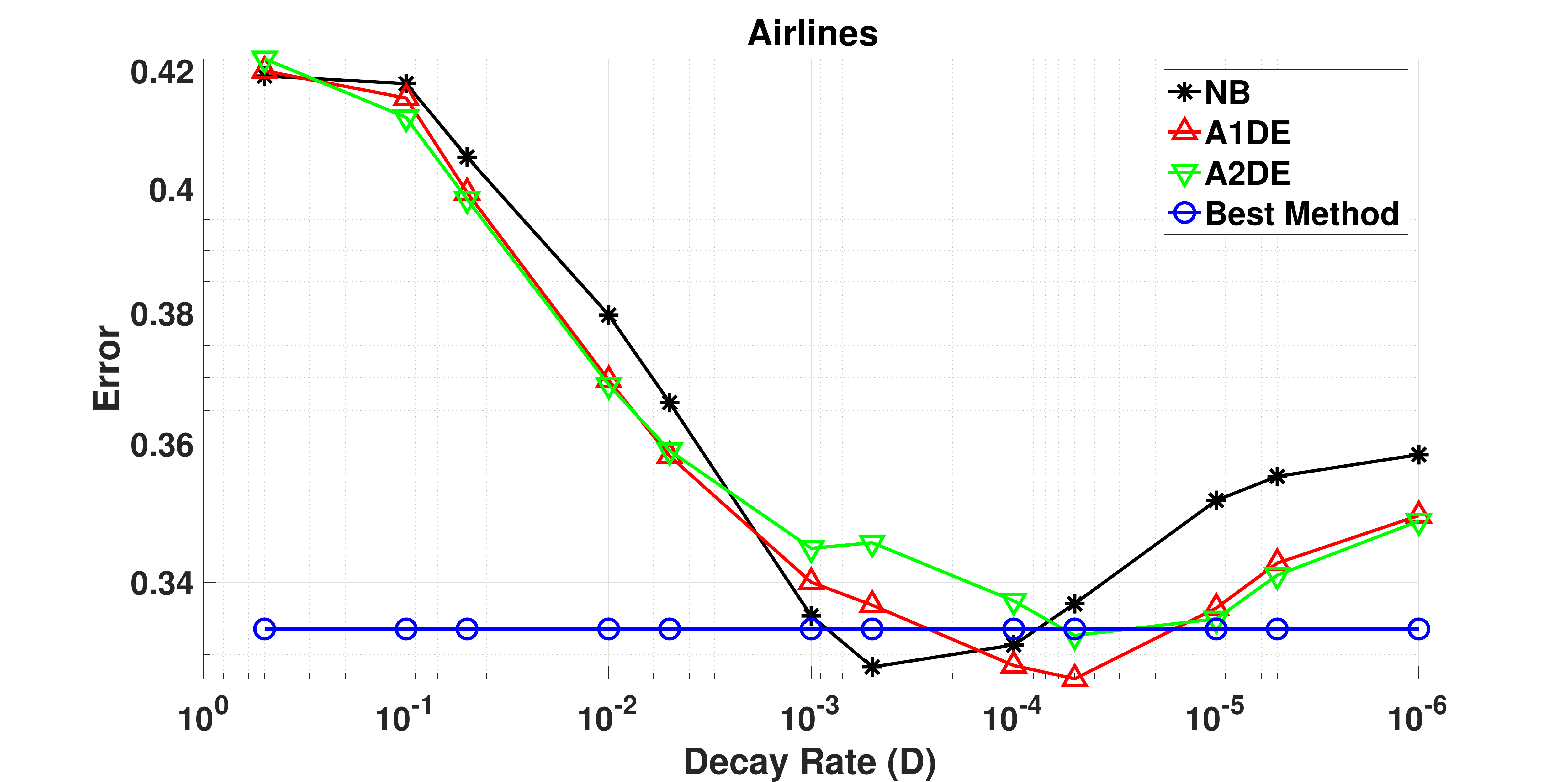}

\hspace{-0.2in}
\includegraphics[width=60mm,height=45mm]{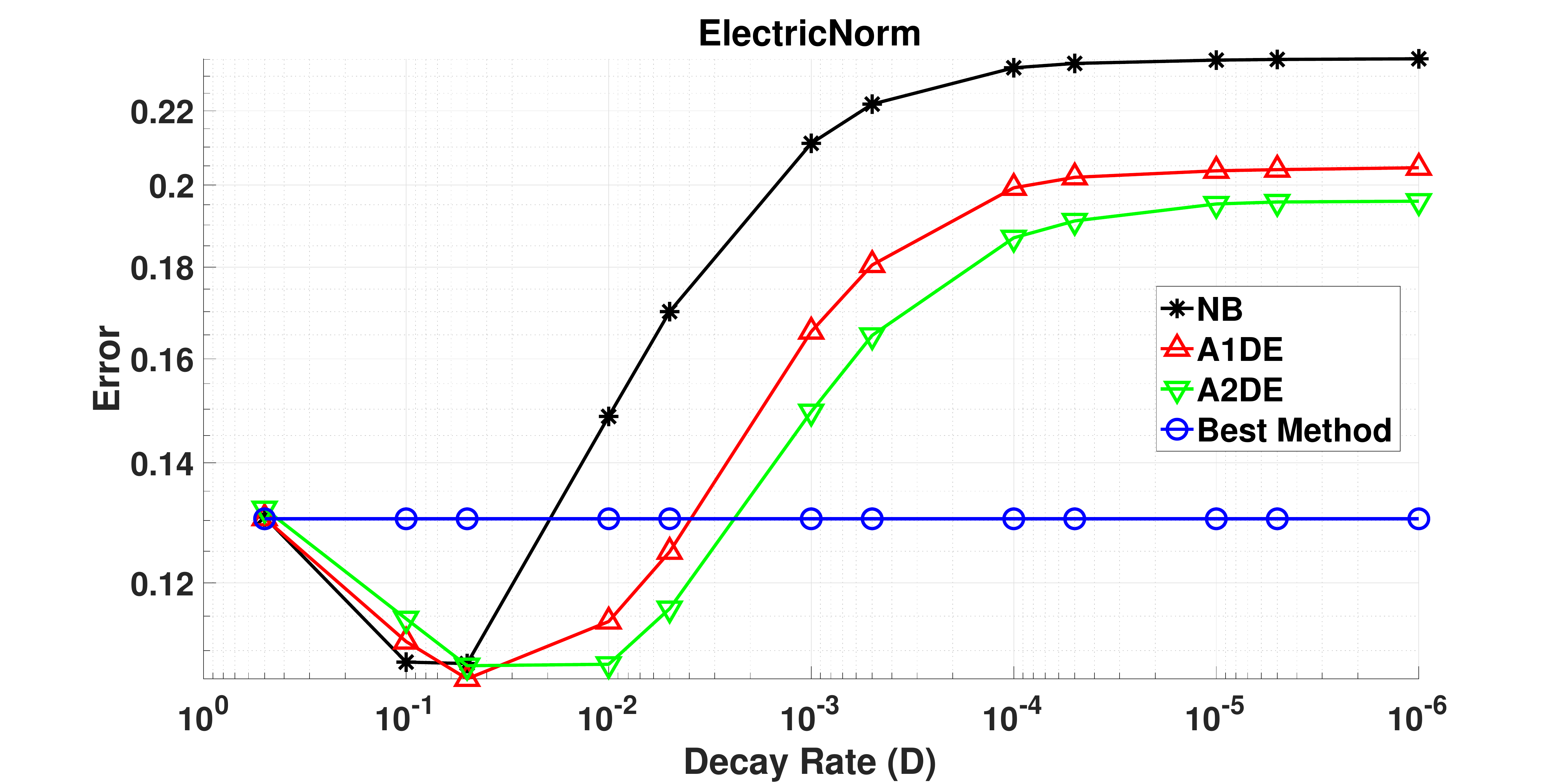}
\includegraphics[width=60mm,height=45mm]{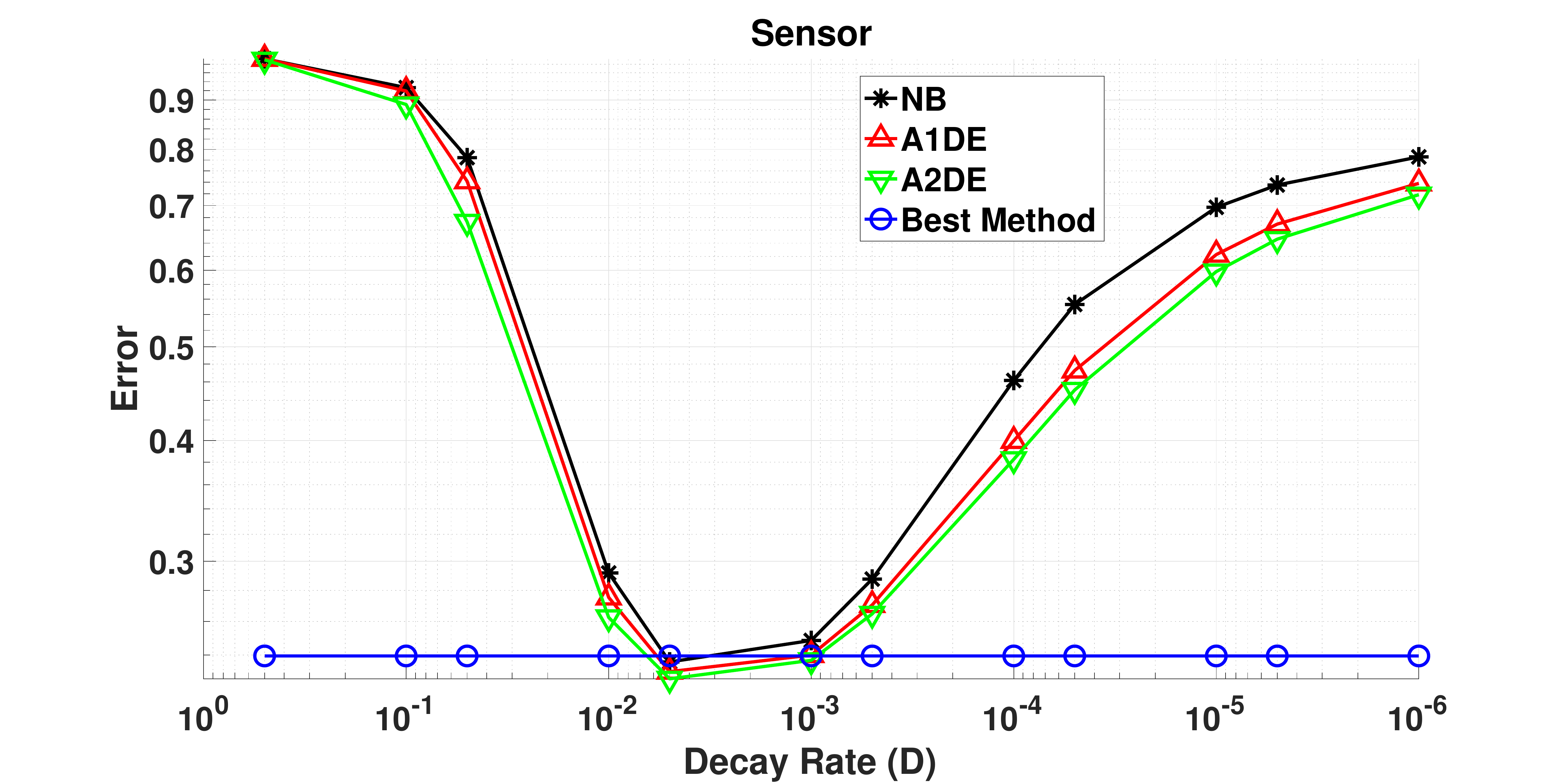}

\vspace*{-5pt}\caption{\small Comparison of adaptive NB, A1DE and A2DE based on decay-based adaptation on four real world datasets: \texttt{PowerSupply, Airlines, ElectricNorm, Sensor}. Horizontal blue line depicts the best performance out of $12$ standard techniques.}

\label{fig_RW_Comp_D}
\end{figure}

A comparison of adaptive AnDE with decay-based adaptation is given in Figure~\ref{fig_RW_Comp_D}. Similar to window-based results, adaptive AnDE achieved lower error than the lowest achieved by any of the state-of-the-art techniques on all but the \texttt{PowerSupply} dataset. On~\texttt{Airlines} and~\texttt{ElectricNorm}, A1DE achieved error of~$0.3267$ and $0.1061$ respectively, and on~\texttt{Sensor}, A2DE achieved error of $0.2268$. 

Note that we are not suggesting that our learners are currently suited for practical use. There is still need to find effective mechanisms to dynamically select forgetting rates and bias/variance profiles. Also, while our learners outperformed the best of the state-of-the-art, many of those learners could also have performed better if a sweep had been performed over their meta-parameters.

\section{Conclusions} \label{sec_conclusions}

We have proposed novel, generalizable and falsifiable hypotheses about the inter-relationships between drift rates, forgetting mechanisms, bias/variance profiles and error.  Our experiments with the AnDE learners have been consistent with the predictions of the hypotheses, that there will be a \emph{sweet path} whereby as the drift rate increases the optimal forgetting rates will also increase; and as forgetting rates increase the optimal model variance profile will decrease.

The AnDE classifiers are well suited to this study due to their efficient support of incremental learning with both windowing and decay and the range of bias/variance profiles that they provide.

Our studies with real-world data show that this framework can result in prequential accuracies that are highly competitive with the best of the state-of-the-art.
Development of practical techniques for dynamically selecting and adjusting forgetting rates and bias/variance profiles as drift rates vary remains a promising avenue for future research.

Another intriguing insight that invites further investigation arises from the observation that for fast drift the lowest bias learner achieved lowest error with the slowest forgetting rate, in apparent disagreement with our hypotheses. As we discuss in Section~\ref{sec_experiments}, this is due to its failure to learn from the fast drifting attributes and hence its performance being dominated by the attribute subspace that is stationary ($Y$, $X_1$ and $X_2$). This gives support to our argument elsewhere \citep{WebbEtAl18} that it is important to analyze drift in marginal distributions as well as at the course global level. It invites the development of learners that bring different forgetting rates and bias/variance profiles to bear on different attribute sub-spaces at different times as their rates of drift vary.

We hope that our hypotheses will provide insights into how to optimize a wide range of mechanisms for handling concept drift and will stimulate future research.

\section{Acknowledgments} \label{sec_ack}

This material is based upon work supported by the Air Force Office of Scientific Research, Asian Office of Aerospace Research and Development (AOARD) under award number FA2386-17-1-4033.

\appendix 

\section{Code} 

The code used in this work can be downloaded from repository:~\url{https://github.com/nayyarzaidi/SweetPathVoyager}.

\bibliography{template.bib}
\bibliographystyle{spbasic}

\end{document}